\documentclass[journal]{IEEEtran}
\usepackage{times}
\usepackage{graphicx}
\usepackage{epsfig}
\usepackage{amsmath}
\usepackage{amssymb}
\usepackage{amsthm}
\usepackage{array}
\usepackage{multirow}
\usepackage{subcaption}
\usepackage{color}
\usepackage{url}
\usepackage{breqn}
\usepackage[font=scriptsize]{caption}
\newtheorem{theorem}{Theorem}
\newtheorem{corollary}{Corollary}

\usepackage{fancyhdr}
\fancypagestyle{firstpage}{%
  \lfoot{\small{The results of diffGrad in this version are improved than earlier IEEE version published at: https://ieeexplore.ieee.org/document/8939562. \copyright2019 IEEE.
Personal use of this material is permitted. Permission from IEEE must be obtained for all other users, including reprinting/republishing this material for advertising or promotional purposes, creating new collective works for resale or redistribution to servers or lists, or reuse of any copyrighted components of this work in other works.}}
}

\begin{document}
\title{diffGrad: An Optimization Method for Convolutional Neural Networks}

\author{
Shiv Ram Dubey, ~\IEEEmembership{Member,~IEEE,}
Soumendu Chakraborty, Swalpa Kumar Roy, ~\IEEEmembership{Student Member,~IEEE,}
Snehasis Mukherjee, ~\IEEEmembership{Member,~IEEE,}
Satish Kumar Singh, ~\IEEEmembership{Senior Member,~IEEE,}
and
Bidyut Baran Chaudhuri, ~\IEEEmembership{Life~Fellow,~IEEE}
\thanks{S.R. Dubey and S. Mukherjee are with the Computer Vision Group, Indian Institute of Information Technology, Sri City, Chittoor, Andhra Pradesh-517646, India (e-mail: shivram1987@gmail.com, srdubey@iiits.in, snehasis.mukherjee@iiits.in). }
\thanks{S. Chakraborty is with the Indian Institute of Information Technology, Lucknow, Uttar Pradesh, India (email: soum.uit@gmail.com).}
\thanks{S.K. Roy and B.B. Chaudhuri are with the Computer Vision and Pattern Recognition Unit at Indian Statistical Institute, Kolkata-700108, India (email: swalparoy@gmail.com, bidyutbaranchaudhuri@gmail.com). Prof. Chaudhuri is also with the Techno India University, Sector V, Salt Lake City, Kolkata-700091, India.}
\thanks{S.K. Singh is with the Computer Vision and Biometrics Laboratory at Indian Institute of Information Technology, Allahabad-211015, India (email: sk.singh@iiita.ac.in).}
}
\markboth{}
 {}

\maketitle
\thispagestyle{firstpage}

\begin{abstract}
Stochastic Gradient Decent (SGD) is one of the core techniques behind the success of deep neural networks. The gradient provides information on the direction in which a function has the steepest rate of change. The main problem with basic SGD is to change by equal sized steps for all parameters, irrespective of gradient behavior. Hence, an efficient way of deep network optimization is to make adaptive step sizes for each parameter. Recently, several attempts have been made to improve gradient descent methods such as AdaGrad, AdaDelta, RMSProp and Adam. These methods rely on the square roots of exponential moving averages of squared past gradients. Thus, these methods do not take advantage of local change in gradients. In this paper, a novel optimizer is proposed based on the difference between the present and the immediate past gradient (i.e., diffGrad). In the proposed diffGrad optimization technique, the step size is adjusted for each parameter in such a way that it should have a larger step size for faster gradient changing parameters and a lower step size for lower gradient changing parameters. The convergence analysis is done using the regret bound approach of online learning framework. Rigorous analysis is made in this paper over three synthetic complex non-convex functions. The image categorization experiments are also conducted over the CIFAR10 and CIFAR100 datasets to observe the performance of diffGrad with respect to the state-of-the-art optimizers such as SGDM, AdaGrad, AdaDelta, RMSProp, AMSGrad, and Adam. The residual unit (ResNet) based Convolutional Neural Networks (CNN) architecture is used in the experiments. The experiments show that diffGrad outperforms other optimizers. Also, we show that diffGrad performs uniformly well for training CNN using different activation functions. The source code is made publicly available at \url{https://github.com/shivram1987/diffGrad}.
\end{abstract}

\section{Introduction}
During the last few years, deep learning based techniques have gained more and more popularity in solving problems in different domains, especially where a data driven approach is required \cite{DeepLearning}. Due to the availability of GPU-based high-end computational facilities and the huge amount of data, deep learning based approaches generally outperform the traditional hand-designed approaches to solve research problems in Computer Vision \cite{AlexNet}, \cite{lan2018learning}, \cite{VggNet}, \cite{shao2019joint}, Image Processing \cite{dong2016image}, \cite{chen2018deeplab}, Signal Processing \cite{yu2011deep}, \cite{zhang2013deep}, Robotics \cite{mnih2015human}, Natural Language Processing \cite{collobert2008unified}, \cite{greff2017lstm}, and many other diverse areas of Artificial Intelligence. Other applications where deep learning can be used include object tracking \cite{lan2015joint}, \cite{zhang2016biologically}, \cite{lan2019learning}, face anti-spoofing and micro-expression recognition \cite{reddy2019spontaneous}, \cite{nagpal2018performance}, hyper-spectral image classification \cite{roy2019hybridsn}, etc. 

\begin{figure*}[t]
  \begin{subfigure}{.33\textwidth}
	\centering
  	\includegraphics[width=.9\linewidth]{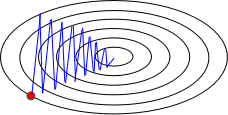}
    \caption{}
    \label{fig:sgd1}
  \end{subfigure}%
  \begin{subfigure}{.33\textwidth}
    \centering
    \includegraphics[width=1\linewidth]{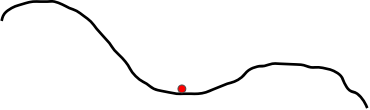}
    \caption{}
    \label{fig:sgd2}
  \end{subfigure}
  \begin{subfigure}{.33\textwidth}
    \centering
    \includegraphics[width=.9\linewidth]{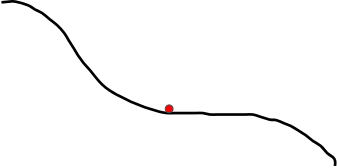}
    \caption{}
    \label{fig:sgd3}
  \end{subfigure}
  \caption{(a) An optimization landscape for two parameters represented in two directions, \emph{i.e.} horizontal and vertical directions. The loss changes slow in horizontal direction and fast in vertical direction. (b) A local minima scenario in 1-D optimization. (c) A saddle point scenario in 1-D optimization.}
  \label{fig:sgd}
\end{figure*}

The deep neural network has different variants to deal with the different problems, such as Convolutional Neural Networks (CNN) and Generative Adversarial Networks (GAN) for images, Recurrent Neural Network (RNN) and Long Short Term Memory Network (LSTM) for temporal sequences of data, 3D-CNN for videos, etc. Research on CNN has observed a rapid growth in recent years, especially on various image based problems. Different CNN architectures have been proposed for image related problems such as AlexNet \cite{AlexNet}, VggNet \cite{VggNet}, GoogLeNet \cite{GoogleNet}, and ResNet \cite{ResNet} for image classification, 
R-CNN \cite{rcnn}, Fast R-CNN \cite{fastrcnn}, 
Faster R-CNN \cite{fasterrcnn}, and YOLO \cite{yolo} for object detection, Mask R-CNN \cite{maskrcnn} and PANet \cite{panet} for instance segmentation, RCCNet \cite{rccnet} for colon cancer nuclei classification, etc. 

In most neural networks, the basic approach that is usually followed for finding an optimal solution is Stochastic Gradient Descent (SGD) optimization \cite{SGD}. Here, initially a measure of loss over the current parameter values is computed using a loss function defined for the specific problem being solved. Then, the gradient for each parameter (i.e. in each dimension) in the network is computed and the parameter values are updated in the opposite direction of the gradient by a factor proportional to the gradient. For SGD optimization, the above two steps are repeated until convergence or until a certain number of epochs or iterations are completed. There are the following four major drawbacks in the basic SGD approach: 1) If the loss changes quickly for a set of parameters and slowly for another set of parameters, then it leads to very slow learning along shallow dimensions and a jittering effect along steep dimensions \cite{sutton1986two} as depicted in Fig. \ref{fig:sgd1}. 2) If the loss function has a local minimum or a saddle point \cite{SaddlePoint}, then SGD gets stuck due to zero gradients in that region. These situations are illustrated in Fig. \ref{fig:sgd2} and Fig. \ref{fig:sgd3}. 3) The gradients are usually computed over minibatches; so the gradients can be noisy. 4) SGD takes the same step for each parameter irrespective of the iteration-wise gradient behavior for that parameter, which leads to poor optimization. Note that each parameter of the network is considered as a dimension in the optimization landscape. Therefore, both dimension and parameter are used interchangeably in this paper. 

Improving SGD based optimization method for neural networks has recently become an active area of research interest. In order to address the above problems, several SGD variants have been proposed in the literature. For example, SGD with momentum (SGDM) is an extension of SGD by incorporating the past gradients in each dimension \cite{qian1999momentum}, \cite{SGDM}. SGDM maintains a momentum in each dimension as a function of previous momentum and current gradient in that dimension. The goal of SGDM is to develop high ``velocity" in any dimension that has a consistent gradient. In SGDM, the jittering problem is reduced with high velocity in consistent gradient dimensions and the saddle point problem is reduced by using the past gradients, which provide some momentum, even when the current gradient is close to zero. SGDM optimization is further improved by Sutskever et al. by Nesterov’s Accelerated Gradient (NAG), which guarantees a better convergence rate than SGD in certain situations \cite{SGDM}. In NAG, the gradients are computed based on an approximation of the next position of the parameters, which are used to update the current moments and parameters. Thus, NAG updates both the moments and gradients, based on the future dimensions.

Another widely used variant of the gradient descent method is AdaGrad \cite{AdaGrad}, proposed to deal with sparse data. AdaGrad performs larger updates for infrequent parameters which leads to smaller magnitude of gradients and smaller updates for frequent parameters which leads to larger magnitude of gradients. Basically, AdaGrad divides the learning rate with the square root of the sum of the squares of the past gradients for all parameters. Apart from dealing with sparse data, AdaGrad is applied to other kinds of problems as well, such as training large-scale neural nets to recognize cats in Youtube videos \cite{dean2012large} and training GloVe word embeddings, as infrequent words require larger updates than frequent ones \cite{pennington2014glove}. But, AdaGrad accumulates the square of gradients which, in turn, may lower the learning rate drastically after some time and kill the learning process. Zeiler extended AdaGrad to AdaDelta by removing the problem of a dying learning rate, which was caused by the monotonically increasing sum of square of gradients \cite{AdaDelta}. AdaDelta also accumulates the square of past gradients, but considers only a few immediate past gradients instead of all past gradients. RMSProp is another attempt to correct the diminishing learning rate of AdaGrad similar to AdaDelta \cite{RMSProp}, i.e. by accumulating the gradient as an exponentially decaying average of squared gradients. The major difference between AdaDelta and RMSProp is that AdaDelta does not use any learning rate \cite{AdaDelta}, whereas RMSProp uses a learning rate \cite{RMSProp}.

One of the recent and popular variants of gradient descent is Adaptive Moment Estimation (Adam). Adam computes adaptive learning rates for each parameter \cite{Adam} by utilizing both first and second moments. Adam accumulates the exponentially decaying average of past gradients similar to SGDM as first moment. It also accumulates the exponentially decaying average of square of past gradients similar to AdaDelta and RMSProp as second moment. The moment can be imagined as a ball rolling down a slope, where Adam behaves like a heavy ball with friction, which thus prefers flat minima in the error surface \cite{heusel2017gans}. It is observed that Adam performs reasonably well in practice as compared to the other adaptive learning-methods. However, Adam does not utilize the change in immediate past gradient information, which is incorporated in the proposed diffGrad method. Very recently, Reddi et al. proposed AMSGrad as an improvement over Adam \cite{AMSGrad}. AMSGrad considers the maximum of past second moment (i.e., ``long-term memory” of past gradients) in the parameter update procedure. By doing so, AMSGrad imposes more friction in order to avoid the overshooting of the minimum. AMSGrad does not change the learning rate based on the recent gradient behavior and does not deal with the saddle point problem, either. Whereas, the proposed diffGrad method controls the learning rate based on the changes in the gradient. Some of the recent works in stochastic gradient methods include the Predictive Local Smoothness based SGD (PLS-SGD) \cite{plssgd}, the sign of each minibatch SGD (signSGD) \cite{signsgd}, and Nostalgic Adam (NosAdam) \cite{nosadam}, etc. 

This paper proposes a difference of gradient based optimizer, which improves the well known Adam \cite{Adam} with the difference of gradients (diffGrad) over the iterations. 
The main contributions of this paper are summarized as follows:
\begin{itemize}
\item This paper proposes a new diffGrad gradient descent optimization method for Convolutional Neural Networks (CNN) by considering the local gradient change information between the current and immediate past iteration.
\item We show how the ``short-term gradient behavior" can be utilized to control the learning rate in the optimization landscape in terms of the optimization stage, i.e., near or far from an optimum solution. If change in gradient is large, it means that the optimization is not stable due to local optima, salient region or other factors, and diffGrad allows a high learning rate. If change in gradient is small, it means that the optimization is likely to be close to the optimum solution, and diffGrad lowers the learning rate automatically.
\item The proposed method also utilizes the accumulation of past gradients over iterations to deal with saddle points.
\item We conduct a convergence analysis of diffGrad in terms of the regret bound using the online learning framework. We also derive a proof for diffGrad convergence.
\item An empirical analysis is done by modeling the optimization problem as a regression problem to show the advantages of the proposed diffGrad optimization method over three synthetic complex non-convex functions.
\item We conduct an experimental study on the proposed method and observe its improved performance for an image categorization task using the ResNet based the CNN architecture. We also experiment with different variants of diffGrad and different activation functions.
\end{itemize}

The rest of the paper is structured in the following manner: Section II presents the preliminaries in SGD optimization; Section III proposes the diffGrad optimization method; Section IV conducts the convergence analysis; Section V is devoted to the empirical analysis; Section VI presents the experimental setup; Section VII presents the experimental results, comparison and analysis; and Section VIII provides the concluding remarks.

\section{Preliminaries}
In SGD, all parameters are updated with the same learning rate \(\alpha_t\) in the $t^{th}$ iteration as 
\begin{equation}
\theta_{t+1,i}=\theta_{t,i}-\alpha_t \times g_{t,i},
\end{equation}
where $\theta_{t,i}$ and $\theta_{t+1,i}$ are the previous and updated values for the $i^{th}$ parameter with $i=1,2,...,d$, where $d$ is the number of parameters, and $g_{t,i}$ is the gradient with respect to the parameter $\theta_{t,i}$ for a loss function $\pounds$, defined as
\begin{equation}
g_{t,i}=\frac{\partial(\pounds_{t,\theta})}{\partial(\theta_{t,i})},
\label{gradient}
\end{equation}
where $\pounds_{t,\theta}$ is a loss function with respect to the parameters of the network ($\theta$) in $t^{th}$ iteration. In this paper, the cross entropy loss used for image categorization experiments is defined as
\begin{equation}
\pounds_{t,\theta}=\frac{1}{N_{b}}\sum_{j=1}^{N_{b}}{\pounds_{t,\theta,j}}+\sigma R_{t,\theta},
\end{equation}
where $N_{b}$ is the number of training images in the batch, $\pounds_{t,\theta,j}$ is the cross entropy data loss for $j^{th}$ training image in $t^{th}$ iteration, $R_{t,\theta}$ is the regularization loss in $t^{th}$ iteration, and $\sigma$ is a regularization loss hyper-parameter. The cross entropy data loss $\pounds_{t,\theta,j}$ for $j^{th}$ training image is computed as
\begin{equation}
\pounds_{t,\theta,j}=-log\left( \frac{e^{S_{o_j}}}{\sum_{k=1}^{N_c}{e^{S_k}}} \right),
\end{equation}
where $N_c$ is the total number of classes in the dataset, $o_j$ is the actual class (i.e., ground truth class) for $j^{th}$ training image and $S_k$ is the computed class score for $k^{th}$ class for $j^{th}$ training image. The regularization loss $R_{t,\theta}$ is computed as
\begin{equation}
R_{t,\theta}=\sum_{i=1}^{d}{(\theta_{t,i})^2}.
\end{equation}

In SGDM \cite{SGDM}, the gradient in each dimension is incorporated to gain moment for the parameters having consistent gradient, as follows:
\begin{equation}
m_{t,i}=\beta m_{t-1,i}+g_{t,i},
\end{equation}
\begin{equation}
\theta_{t+1,i}=\theta_{t,i}-\alpha m_{t,i},
\end{equation}
where $m_{t,i}$ is the moment gained at $t^{th}$ iteration for $i^{th}$ parameter $\theta_{t,i}$ with $m_{t,i}=0$ for $t=0$ and $\beta$ is a hyper-parameter to control the moment.

In AdaGrad \cite{AdaGrad}, the basic SGD approach is modified by normalizing the learning rate $\alpha_t$ as 
\begin{equation}
\theta_{t+1,i}=\theta_{t,i}-\frac{\alpha_t \times g_{t,i}}{\sqrt[]{G_{t,i}+\epsilon}},
\end{equation} 
where $\epsilon$ is a small value to avoid division by zero and $G_{t,i}$ is the sum of the squares of the gradients of $t$ steps for the $i^{th}$ parameter and given as
\begin{equation}
G_{t,i}=\sum_{t=1}^{t} (g_{t,i})^2,
\end{equation} 
where $g_{t,i}$ is given by Eq. (\ref{gradient}).
Over the iterations, the value of $G_{t,i}$ may become very large due to the positive accumulation of the square of the gradients and may decrease the effective learning rate $\alpha$ drastically, which in turn can kill the learning process. This problem has been addressed in AdaDelta \cite{AdaDelta} and RMSProp \cite{RMSProp} by leaking the accumulated square of gradients $G_{t,i}$ with a decay rate $\beta$. The $G_{t,i}$ in RMSProp is modified as 
\begin{equation}
G_{t,i}= \beta G_{t-1,i}+(1-\beta)(g_{t,i})^2
\end{equation}
where $G_{t-1,i}=0$ for $t=1$.

Adam \cite{Adam} is another widely used gradient descent optimization technique that computes the learning rates at each step based on two vectors known as the $1^{st}$ and $2^{nd}$ order moments (i.e., mean and variance, respectively), which are recursively defined using the gradient and the square of the gradient, respectively. Basically, Adam is an improvement over RMSProp by incorporating first moment with RMSProp. Here, the $1^{st}$ and $2^{nd}$ order moments are defined as
\begin{equation}
m_{t,i}= \beta_1 m_{t-1,i}+\left(1-\beta_1\right)g_{t,i}
\label{first_moment}
\end{equation}
\begin{equation}
v_{t,i}= \beta_2 v_{t-1,i}+\left(1-\beta_2\right)g_{t,i}^2
\label{second_moment}
\end{equation}
where $\beta_1$ and $\beta_2$ are the decay rates for first and second moments, respectively, $m_{t-1,i}$ and $v_{t-1,i}$ are the mean and variance of the gradient of the previous steps, respectively. Both $m_{t-1,i}$ and $v_{t-1,i}$ are initialized with $0$ at the $1^{st}$ iteration, $t=1$. It is observed that, initially, the value of first moment is small and the value of second moment is very small, thus leading to a very large step size. In order to solve this issue, Adam has incorporated a bias correction of the $1^{st}$ and $2^{nd}$ order moments as
\begin{equation}
\hat{m}_{t,i}=\frac{m_{t,i}}{(1-\beta_1^t)} \hspace{5mm}\text{and}\hspace{5mm} \hat{v}_{t,i}=\frac{v_{t,i}}{(1-\beta_2^t)},
\label{biased_correction}
\end{equation}
where $\beta_1^t$ is $\beta_1$ power $t$, $\beta_2^t$ is $\beta_2$ power $t$, and $\hat{m}_{t,i}$ and $\hat{v}_{t,i}$ are the biased corrected first and second moments, respectively. Thus, the parameter update in Adam is incorporated as
\begin{equation}
\theta_{t+1,i}=\theta_{t,i}-\frac{\alpha_t \times \hat{m}_{t,i}}{\sqrt[]{\hat{v}_{t,i}+\epsilon}}.
\end{equation}

A problem arises in Adam when the value of the second moment deceases significantly. Due to low values of the second moment, the friction in the optimization landscape decreases, which leads to the situation where the update process overshoots an optimum solution due to a high learning rate and diverges. This problem is addressed in AMSGrad \cite{AMSGrad} by considering the maximum of second moments in current and past iterations. In AMSGrad, the $1^{st}$ and $2^{nd}$ order moments are computed and bias-corrected similar to Adam. However, AMSGrad normalizes the learning rate $\alpha_t$ by the maximum of all $\hat{v}^{max}_{t,i}$ values, instead of only $\hat{v}_{t,i}$. AMSGrad memorizes the highest of $2^{nd}$ order moment to give more priority to those steps which update the parameter in a more accurate direction. Thus, in AMSGrad, $\hat{v}^{max}_{t,i}$ is defined as
\begin{equation}
\hat{v}^{max}_{t,i}=max(\hat{v}^{max}_{t-1,i}, \hat{v}_{t,i}),
\end{equation}
where $max$ is the maximum operator and $\hat{v}^{max}_{t-1,i}=0$ for $t=1$. Thus, the parameter update in AMSGrad is carried out using the following update rule:
\begin{equation}
\theta_{t+1,i}=\theta_{t,i}-\frac{\alpha_t \times \hat{m}_{t,i}}{\sqrt[]{\hat{v}^{max}_{t,i}+\epsilon}}.
\end{equation}

In practice, Adam is popular in various problems related to deep learning. Adam with $1^{st}$ order moment hyper-parameter $\beta_1 = 0.9$, $2^{nd}$ order moment hyper-parameter $\beta_2 = 0.999$, and learning rate $\alpha \in [10^{-2}, 10^{-4}]$ is a good starting choice for many models \cite{Adam}.

\begin{figure}[t]
\centering
\includegraphics[width=0.9\linewidth]{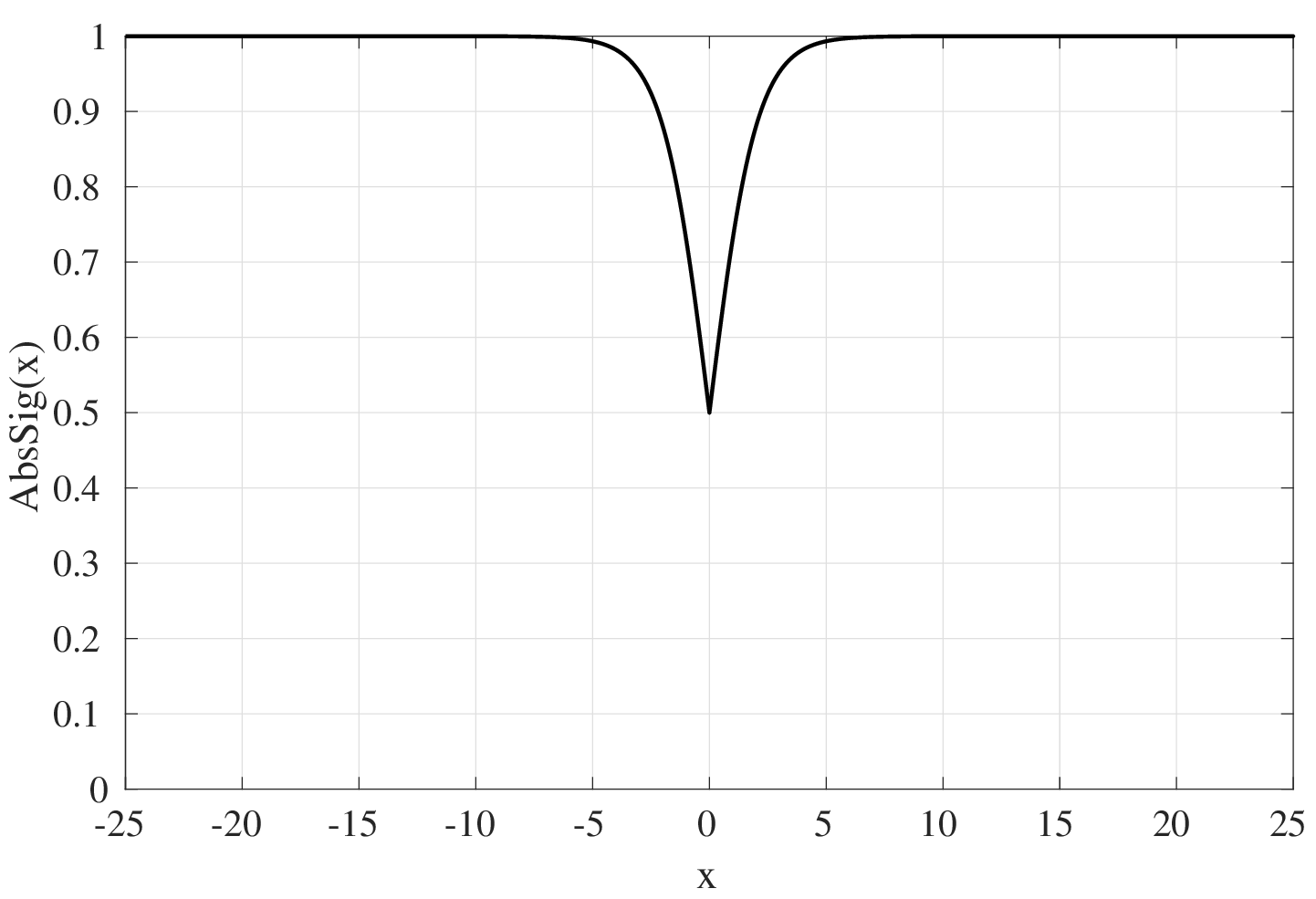}
\caption{The behavior of diffGrad Friction Coefficient (DFC) in terms of the $AbsSig(x)$ values. Here a DFC value of 1 means no friction and 0 means infinite friction. Note that the friction is roughly negligible if the difference in gradient is high. Here, negligible represents DFC$\approx$1 which means there is no friction and in this case diffGrad is the same as Adam. Moreover, the minimum DFC is 0.5 when there is no gradient change. Otherwise, the DFC, close to zero, will slow down the learning process heavily.}
\label{fig:sigmoid}
\end{figure}

\section{Proposed diffGrad Optimization}
From the discussions in the previous section we can conclude that recent optimization techniques such as Adam and AMSGrad suffer from the problem of automatic adjustment of the learning rate. The main problem is with controlling friction for the first moment in order to avoid over shooting near to an optimum solution. In this section, we propose a new gradient descent optimization technique referred to as diffGrad to address these issues of existing gradient descent optimization techniques.
The proposed diffGrad optimization technique is based on the change in short-term gradients and controls the learning rate based on the need of dynamic adjustment of learning rate. This means that diffGrad follows the norm that the parameter update should be smaller in low gradient changing regions and vice-versa. 
diffGrad computes the $1^{st}$ and $2^{nd}$ order moments (i.e., $m_{t,i}$ and $v_{t,i}$, respectively) as well as the $1^{st}$ and $2^{nd}$ order bias-corrected moments (i.e., $\hat{m}_{t,i}$ and $\hat{v}_{t,i}$, respectively) for the $i^{th}$ parameter at the $t^{th}$ iteration similar to Adam \cite{Adam} using Eq. (\ref{first_moment}-\ref{biased_correction}). 

A diffGrad friction coefficient (DFC) is introduced in the proposed work to control the learning rate using information of short-term gradient behavior. The DFC is represented by $\xi$ and defined as
\begin{equation}
\label{absig}
\xi_{t,i}=AbsSig(\Delta g_{t,i})
\end{equation}
where $AbsSig$ is a non-linear sigmoid function that squashes every value between 0.5 and 1, and is defined as
\begin{equation}
AbsSig(x)=\frac{1}{1+e^{-|x|}}
\end{equation}
while $\Delta g_{t,i}$ is the change in gradient between immediate past and current iterations, given as
\begin{equation}
\Delta g_{t,i}=g_{t-1,i}-g_{t,i}
\end{equation}
where $g_{t,i}$ is the computed gradient for the $i^{th}$ parameter at the $t^{th}$ iteration and defined in Eq. (\ref{gradient}).

The behavior of DFC (i.e., $\xi$) is characterized in Fig. \ref{fig:sigmoid} in terms of $AbsSig(x)$ which represents the friction with respect to the change in gradient. It can be observed from Fig. \ref{fig:sigmoid} that large changes in the gradient incur less friction, whereas small changes in the gradient incur more friction with at most 0.5 when there is no change in the gradient. Note that $|\Delta g_{t,i}| \in \mathbb{R}_0^+$, $\forall i\in[1,d]$, and $i\in \mathbb{I}^+$ leads to $\xi_{t,i}\in[0.5,1]$ at any iteration $t$. The DFC imposes more friction when gradient changes slowly and vice-versa.

In the proposed diffGrad optimization method, the steps up to the computation of bias-corrected $1^{st}$ order moment $\hat{m}_{t,i}$ and bias-corrected $2^{nd}$ order moment $\hat{v}_{t,i}$ are the same as those of Adam optimization \cite{Adam}. The diffGrad optimization method updates the $i^{th}$ parameter at the $t^{th}$ iteration using the following update rule:
\begin{equation}
\label{diffgrad_update_rule}
\theta_{t+1,i}=\theta_{t,i}-\frac{\alpha_t\times\xi_{t,i}\times\hat{m}_{t,i}}{\sqrt[]{\hat{v}_{t,i}}+\epsilon},
\end{equation}
where $\theta_{t,i}$ is the weight for $i^{th}$ parameter at $t^{th}$ iteration, $\alpha_t$ is the learning rate for $t^{th}$ iteration, $\epsilon$ is a very small value of approximately $10^{-7}$ added to avoid division by zero, $\xi_{t,i}$ is the diffGrad friction coefficient, and  $\hat{m}_{t,i}$ and $\hat{v}_{t,i}$ are the bias-corrected $1^{st}$ order moment and the bias-corrected $2^{nd}$ order moment, respectively, as defined in Equation (\ref{biased_correction}). 

\begin{figure}[t]
\centering
\includegraphics[width=\linewidth]{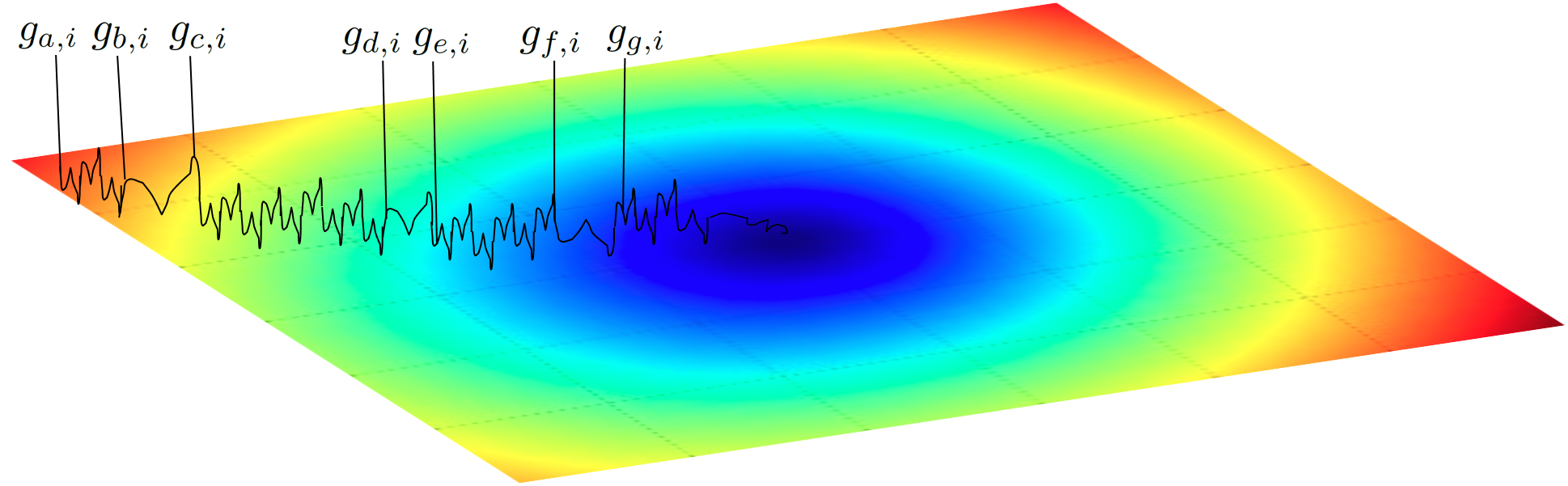}
\caption{The illustration of optimization landscape in order to understand the importance of short-term gradient change. The dark blue area represents the optimization goal to be reached.}
\label{fig:diffgrad}
\end{figure}

The proposed optimizer introduces the diffGrad friction coefficient (DFC) to control the oscillations.
The difference in gradient reduces the learning rate by controlling the moving average near an optimum solution.
The necessity of the friction coefficient is illustrated using an optimization landscape in Fig. \ref{fig:diffgrad}. The dark blue area shows the optimization goal. As we move from dark red to dark blue, the depth of the optima tends to be reduced. Lighter shades between two colors show steep descent towards the optimization goal. Here, $g_{a,i} \rightarrow g_{g,i}$ shows the gradients at steps $a$ to $g$. The color between $g_{a,i}$ and $g_{b,i}$ changes very rapidly. As a result, the learning rate increases due to the increase in the diffGrad friction coefficient. Hence, the frequency of vertical fluctuations also increases. There is a very steep descent between the pairs $g_{c,i} \rightarrow{} g_{d,i}$ and $g_{e,i} \rightarrow g_{f,i}$. Hence, we again see the frequent fluctuations in the vertical direction. The descents between $g_{b,i} \rightarrow g_{c,i}$ and $g_{f,i} \rightarrow g_{g,i}$ are very slow, which is evident from slow or no change in color. Here, the DFC reduces the frequency of fluctuations by decreasing the rate of learning of the network. Hence, the introduction of DFC reduces redundant learning and increases the rate of convergence. It also helps in finding an optimum solution by reducing the vertical fluctuation near local optima.

The problem of an ever decreasing learning rate in Adagrad \cite{AdaGrad} has been solved by Adam \cite{Adam} by introducing two moments on gradient. The adaptive nature of these two moments are controlled during learning with respect to the slope of the descent of the probable solution towards a local optimum. However, a sudden decrease in second moment adversely affects the Adam optimizer and the actual solution overshoots the local optima. This problem has been resolved in AMSGrad \cite{AMSGrad} by introducing a new parameter, $\hat{v}^{max}_{t,i}=max(\hat{v}^{max}_{t-1,i}, \hat{v}_{t,i})$, which prevents the learning rate from overshooting. Adam and AMSgrad ignore the effect of the $1^{st}$ moment in controlling the learning rate over the optimization landscape. The proposed diffGrad optimizer resolves this drawback by introducing a new parameter DFC ($\xi$), which not only allows the high learning rate for a high gradient changing surface, but also reduces the learning rate for a low gradient changing surface and prevents the probable solution from overshooting. The improved performance of diffGrad shows that the optimizer effectively introduces the required friction ($\uparrow$ or $\downarrow$). Please note that the minimum DFC is considered as 0.5, as shown in Fig. \ref{fig:sigmoid}, such that the optimization should not become stuck in local optima and saddle regions. Due to the high moment gained, sufficient magnitude of the step size will be allowed by the DFC in the proposed diffGrad approach, so that it emerges from flat local optima and flat saddle regions.

\section{Convergence Analysis}
The convergence property of Adam \cite{Adam} is shown using the online learning framework proposed in \cite{zinkevich2003online}. We also use this technique to analyze the convergence of the proposed diffGrad optimizer. Consider $f_1(\theta)$, $f_2(\theta)$,$...$, $f_T(\theta)$ as the unknown sequence of convex cost functions. Our aim is to predict parameter $\theta_t$ at each iteration $t$ and evaluate over $f_t(\theta)$. For this type of problem, where the nature of the sequence is not known a priori, the algorithm can be evaluated based on the regret bound. The regret bound is computed by summing the difference between all the previous online guesses $f_t(\theta_t)$ and the best fixed point parameter $f_t(\theta^*)$ from a feasible set $\chi$ for all the previous iterations. Mathematically, the regret bound is given as
\begin{equation}
R(T)=\sum_{t=1}^{T}{[f_t(\theta_t)-f_t(\theta^*)]}
\end{equation}
where $\theta^*=\mbox{arg }\mbox{min}_{\theta \in \chi }\sum_{t=1}^{T}{f_t(\theta)}$. We observe that diffGrad has an $O(\sqrt{T})$ regret bound. The proof is given in the appendix. Our regret bound is comparable to general convex online learning methods. We have used the following definitions: $g_{t,i}$ refers to the gradient in the $t^{th}$ iteration for the $i^{th}$ element, $g_{1:t,i}=[g_{1,i},g_{2,i},...,g_{t,i}] \in \mathbb{R}^t$ is the vector of gradients in the $i^{th}$ dimension over all iterations up to $t$, and $\gamma \triangleq \frac{\beta_1^2}{\sqrt{\beta_2}}$. 

\begin{theorem}
\textit{Consider the bounded gradients for function $f_t$ (i.e., $||g_{t,\theta}||_2 \leq G$ and $||g_{t,\theta}||_{\infty} \leq G_{\infty}$) for all $\theta \in R^d$. Also assume that diffGrad produces the bounded distance between any $\theta_t$ (i.e., $||\theta_n-\theta_m||_2 \leq D$ and $||\theta_n-\theta_m||_\infty \leq D_\infty$ for any $m,n\in\{1,...,T\}$). Let $\gamma \triangleq \frac{\beta_1^2}{\sqrt{\beta_2}}$, $\beta_1,\beta_2 \in [0,1)$ satisfy $\frac{\beta_1^2}{\sqrt{\beta_2}} < 1$, $\alpha_t=\frac{\alpha}{\sqrt{t}}$, and $\beta_{1,t}=\beta_1\lambda^{t-1},\lambda \in (0,1)$ where $\lambda$ is typically very close to $1$, e.g., $1-10^{-8}$. For all $T \geq 1$, the proposed diffGrad optimizer shows the following guarantee:} 
\begin{equation}
\begin{split}
R(T) & \leq \frac{D^2}{2\alpha(1-\beta_1)}\sum_{i=1}^{d}{(1+e^{-|g_{1,i}|})\sqrt{T\hat{v}_{T,i}}} 
\\&+ \frac{\alpha(1+\beta_1) G_\infty}{(1-\beta_1)\sqrt{1-\beta_2}(1-\gamma)^2}\sum_{i=1}^{d}{||g_{1:T,i}||_2} 
\\&+ \sum_{i=1}^{d}{\frac{D_{\infty}^{2}G_{\infty}\sqrt{1-\beta_2}}{2\alpha (1-\beta_1)(1-\lambda)^2}}
\end{split}
\end{equation}
\end{theorem}
Note that the additive term over the dimension ($d$) can be much smaller than its upper bound $\sum_{i=1}^{d}{||g_{1:T,i}||_2}<< dG_\infty\sqrt{T}$ and $\sum_{i=1}^{d}{(1+e^{-|g_{1,i}|})\sqrt{T\hat{v}_{T,i}}} << d(1+E_\infty)G_\infty\sqrt{T}$, where $E_\infty$ is the upper bound over the exponential function and $E_\infty >> \sum_{i=1}^{d}{e^{-|g_{1,i}|}}$. In general, $O(\log d\sqrt{T})$ is achieved by adaptive methods such as diffGrad and Adam which is improved over the $O(\sqrt{dT})$ of non-adaptive methods. The proposed diffGrad method also uses the decay of $\beta_{1,t}$ for the theoretical analysis, similar to Adam.

Finally, by using the above theorem and $\sum_{i=1}^{d}{||g_{1:T,i}||_2}<< dG_\infty\sqrt{T}$ and $\sum_{i=1}^{d}{(1+e^{-|g_{1,i}|})\sqrt{T\hat{v}_{T,i}}} << d(1+E_\infty)G_\infty\sqrt{T}$, the convergence of average regret of diffGrad can be shown as described in the following corollary.
\begin{corollary}
\textit{Consider the bounded gradients for function $f_t$ (i.e., $||g_{t,\theta}||_2 \leq G$ and $||g_{t,\theta}||_{\infty} \leq G_{\infty}$) for all $\theta \in R^d$. Also, assume that diffGrad produces the bounded distance between any $\theta_t$ (i.e., $||\theta_n-\theta_m||_2 \leq D$ and $||\theta_n-\theta_m||_\infty \leq D_\infty$ for any $m,n\in\{1,...,T\}$). For all $T \geq 1$, the proposed diffGrad optimizer shows the following guarantee:} 
\begin{equation}
\frac{R(T)}{T}=O(\frac{1}{\sqrt{T}}). 
\end{equation}
Thus, $\lim_{T\rightarrow\infty}\frac{R(T)}{T}=0$.
\end{corollary}

\begin{figure*}
\centering
  \begin{subfigure}{.3\textwidth}
    \centering
    \includegraphics[width=1\linewidth]{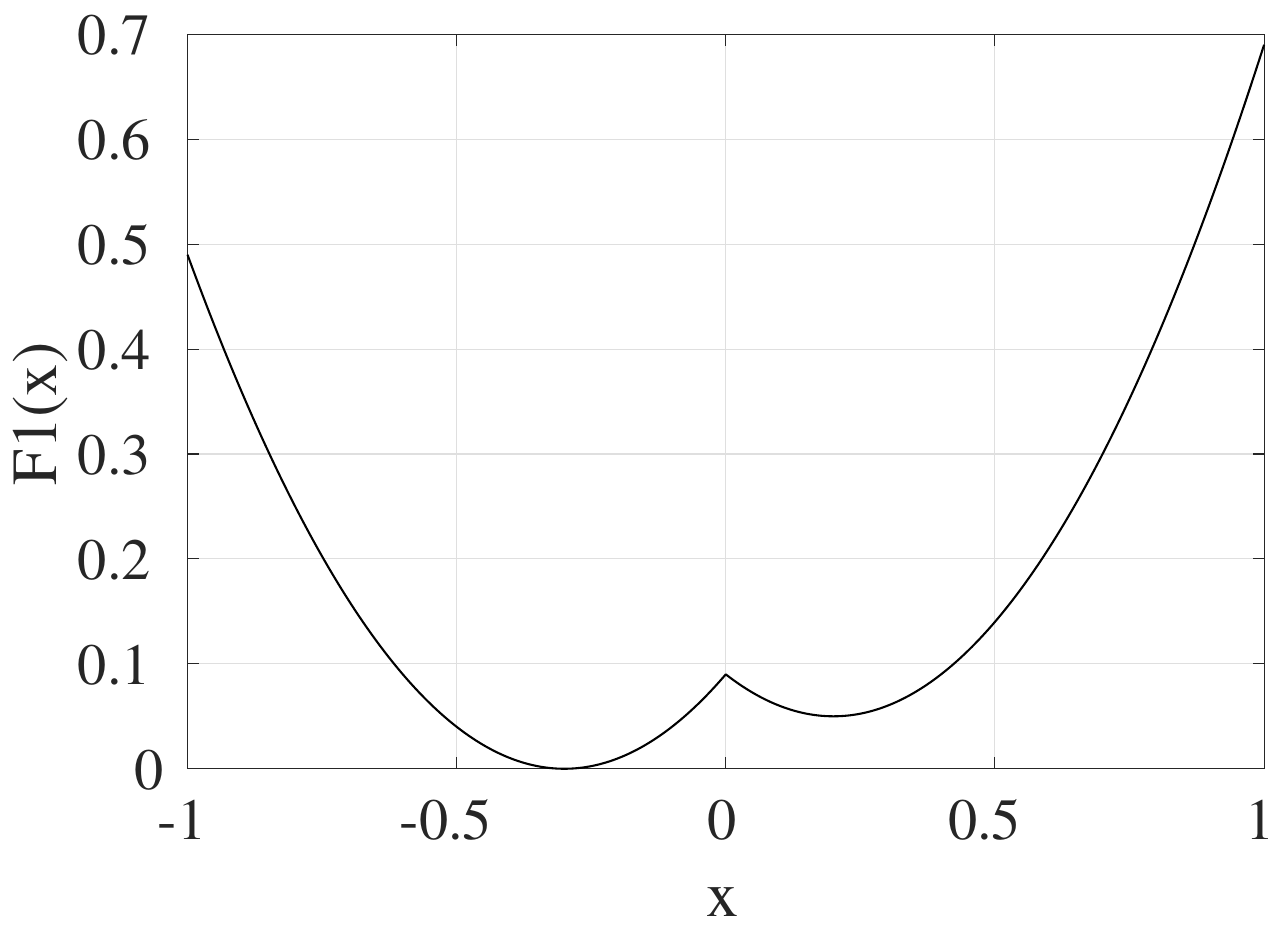}
    \caption{}
    \label{f1}
  \end{subfigure}%
  \begin{subfigure}{.3\textwidth}
    \centering
    \includegraphics[width=1\linewidth]{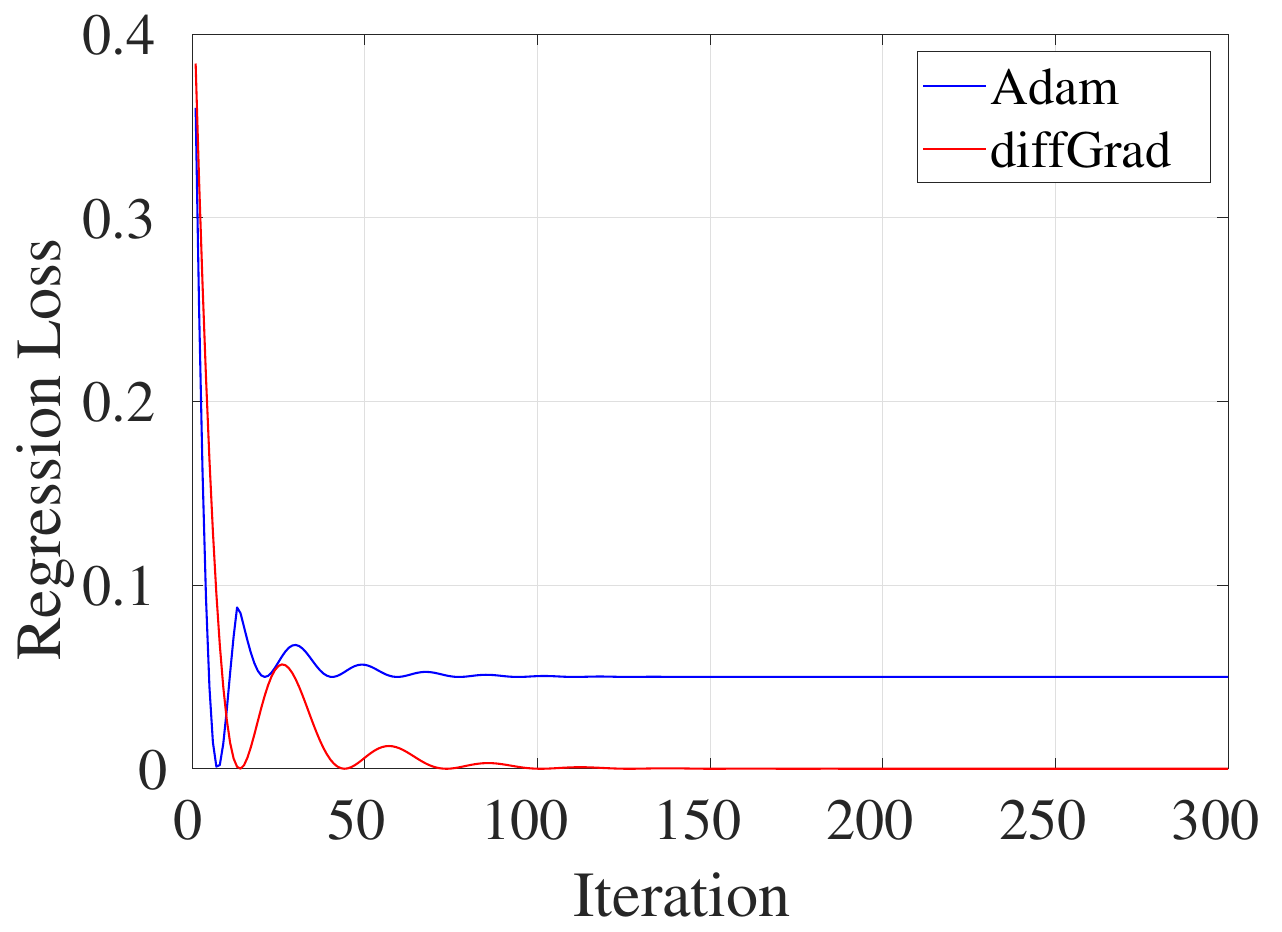}
    \caption{}
    \label{f1_loss}
  \end{subfigure}%
  \begin{subfigure}{.3\textwidth}
    \centering
    \includegraphics[width=1\linewidth]{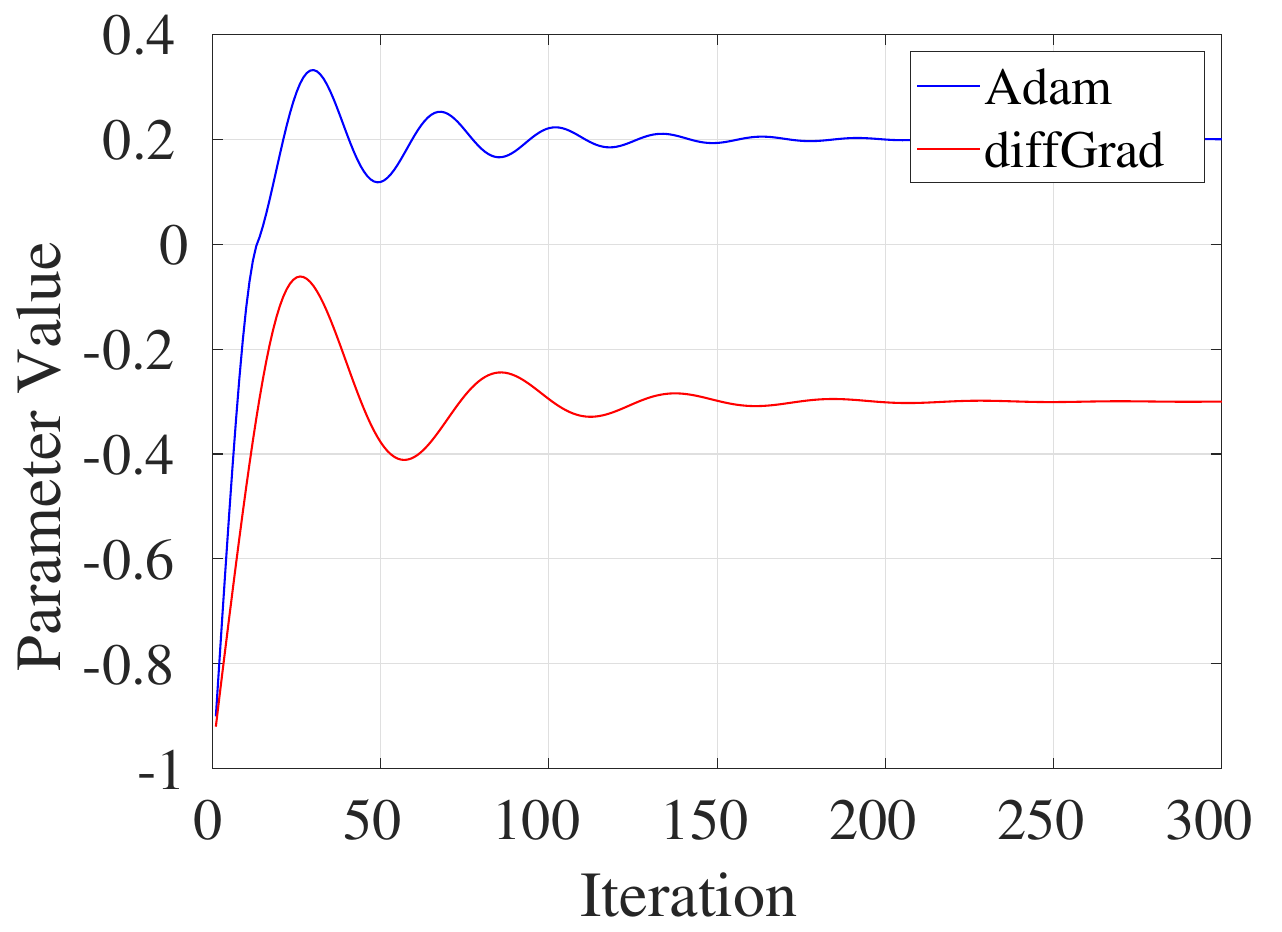}
    \caption{}
    \label{f1_p}
  \end{subfigure}
  \begin{subfigure}{.3\textwidth}
    \centering
    \includegraphics[width=1\linewidth]{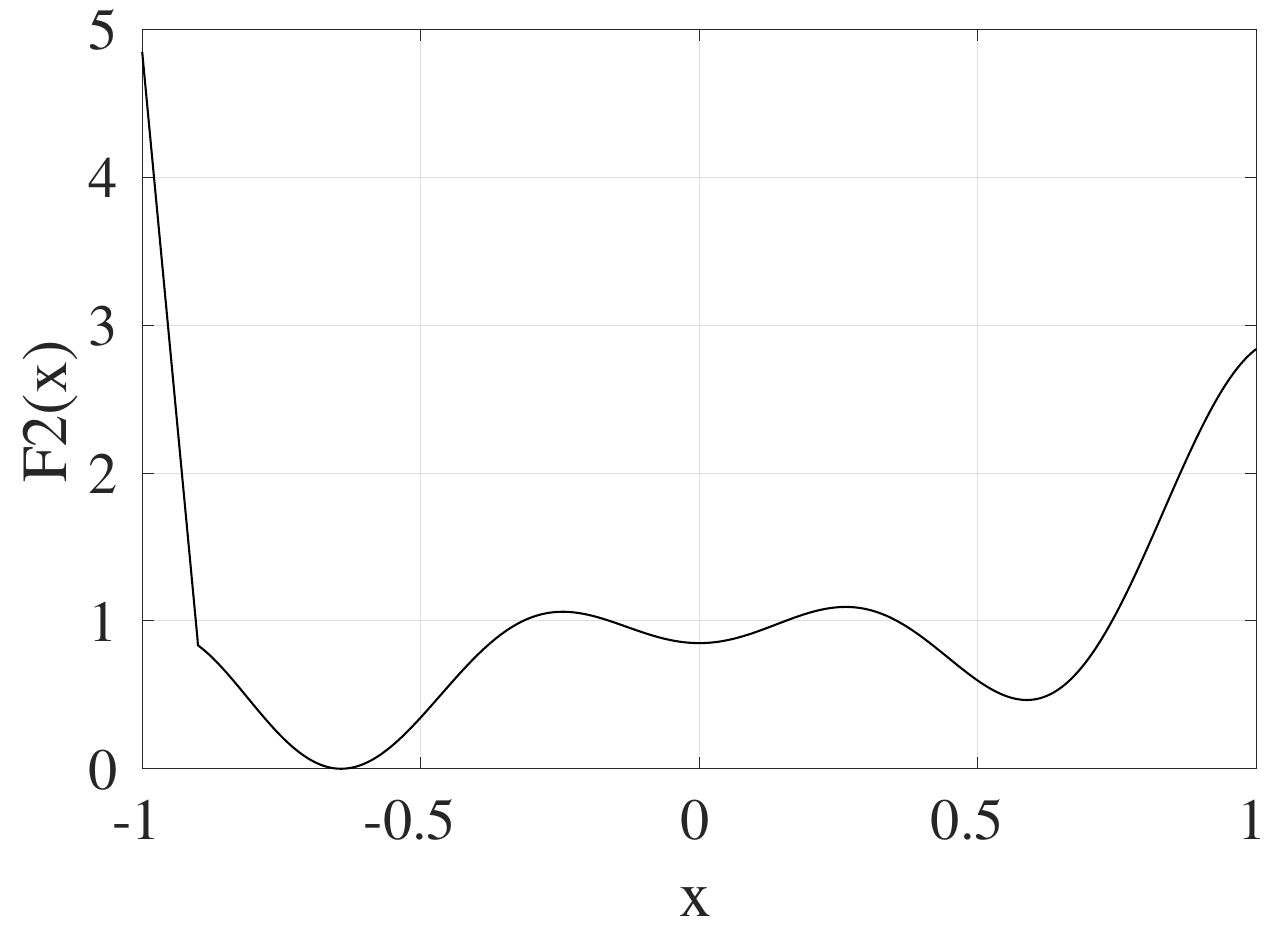}
    \caption{}
    \label{f2}
  \end{subfigure}%
  \begin{subfigure}{.3\textwidth}
    \centering
    \includegraphics[width=1\linewidth]{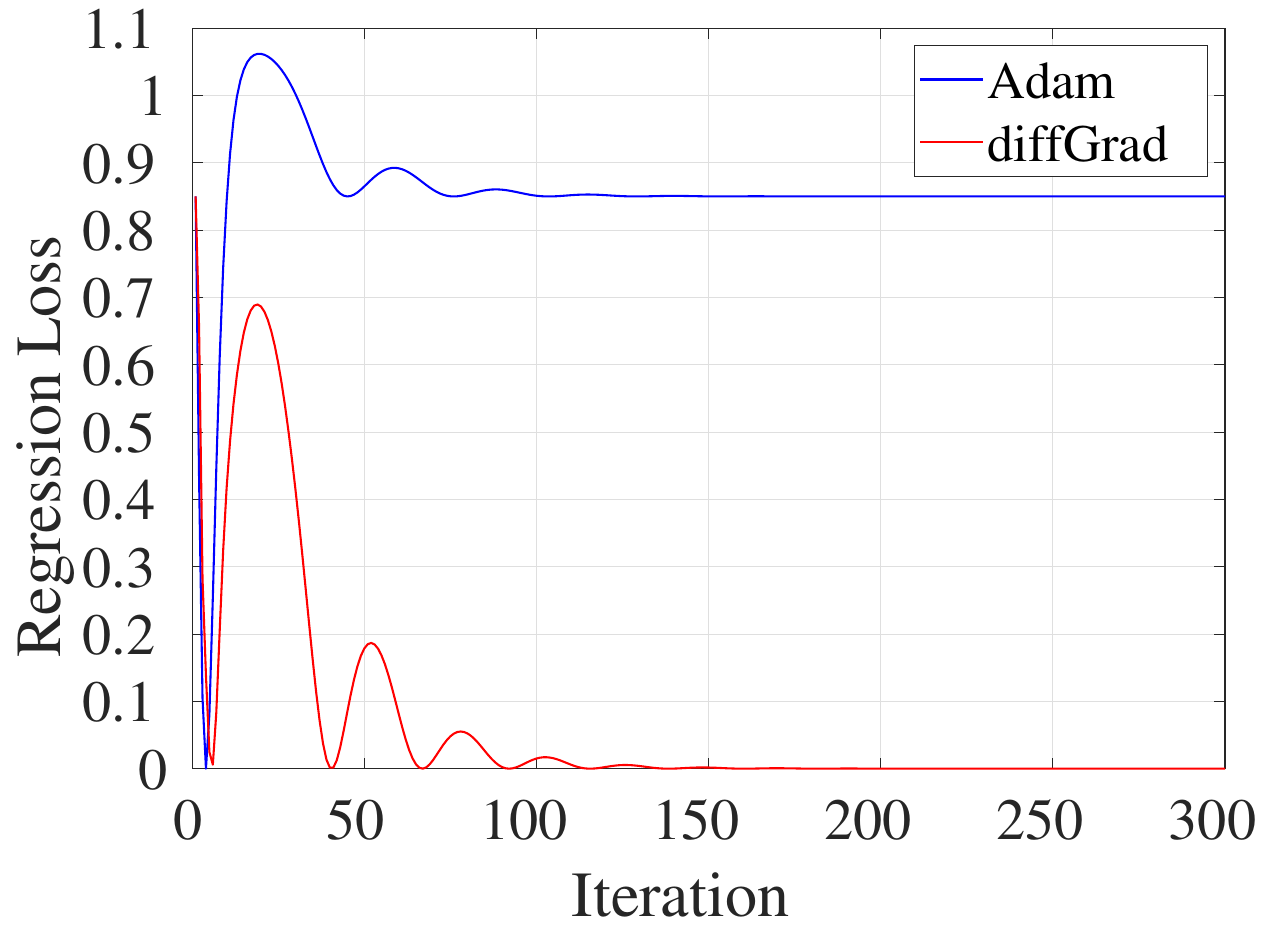}
    \caption{}
    \label{f2_loss}
  \end{subfigure}%
  \begin{subfigure}{.3\textwidth}
    \centering
    \includegraphics[width=1\linewidth]{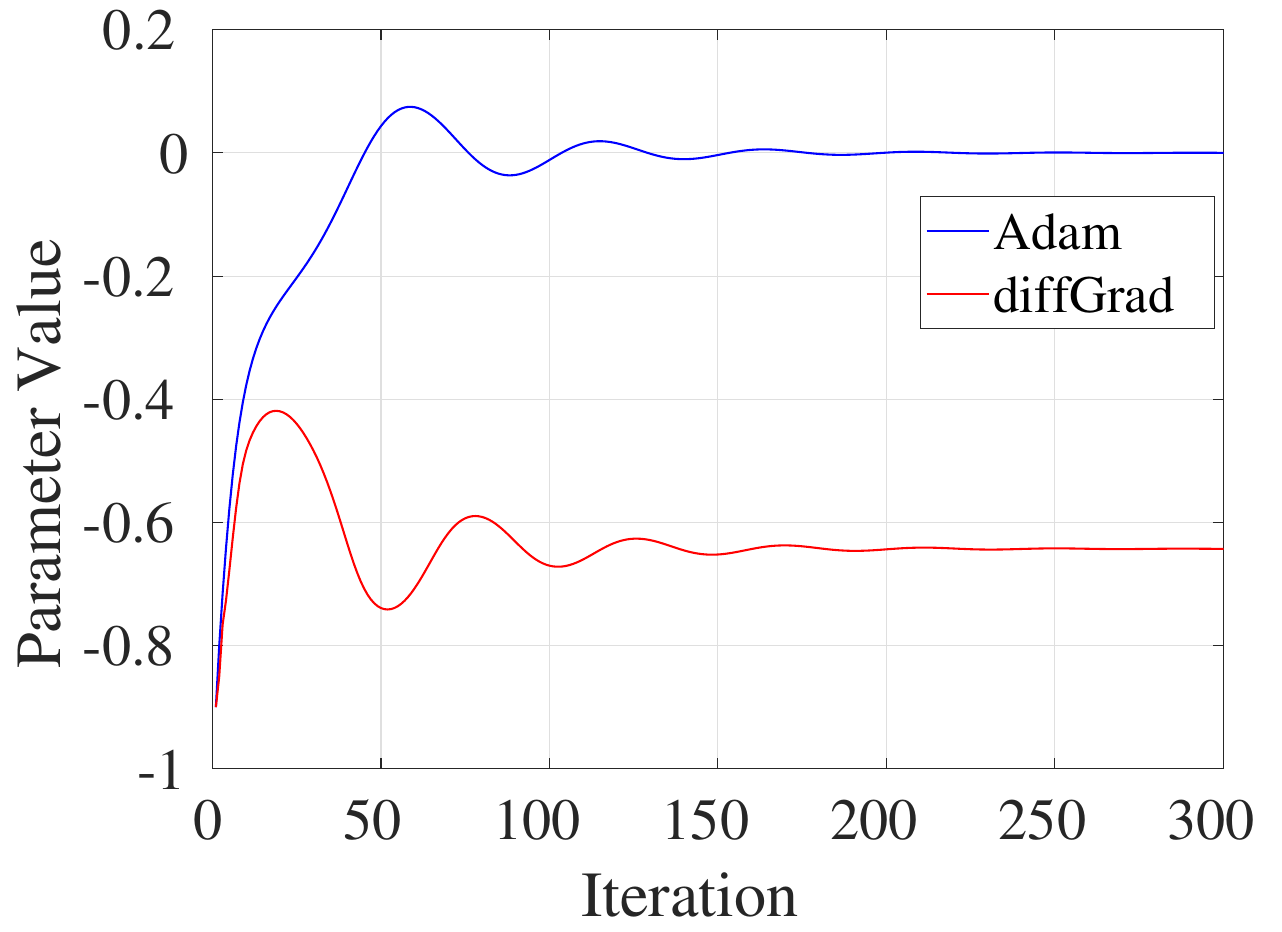}
    \caption{}
    \label{f2_p}
  \end{subfigure}
  \begin{subfigure}{.3\textwidth}
    \centering
    \includegraphics[width=1\linewidth]{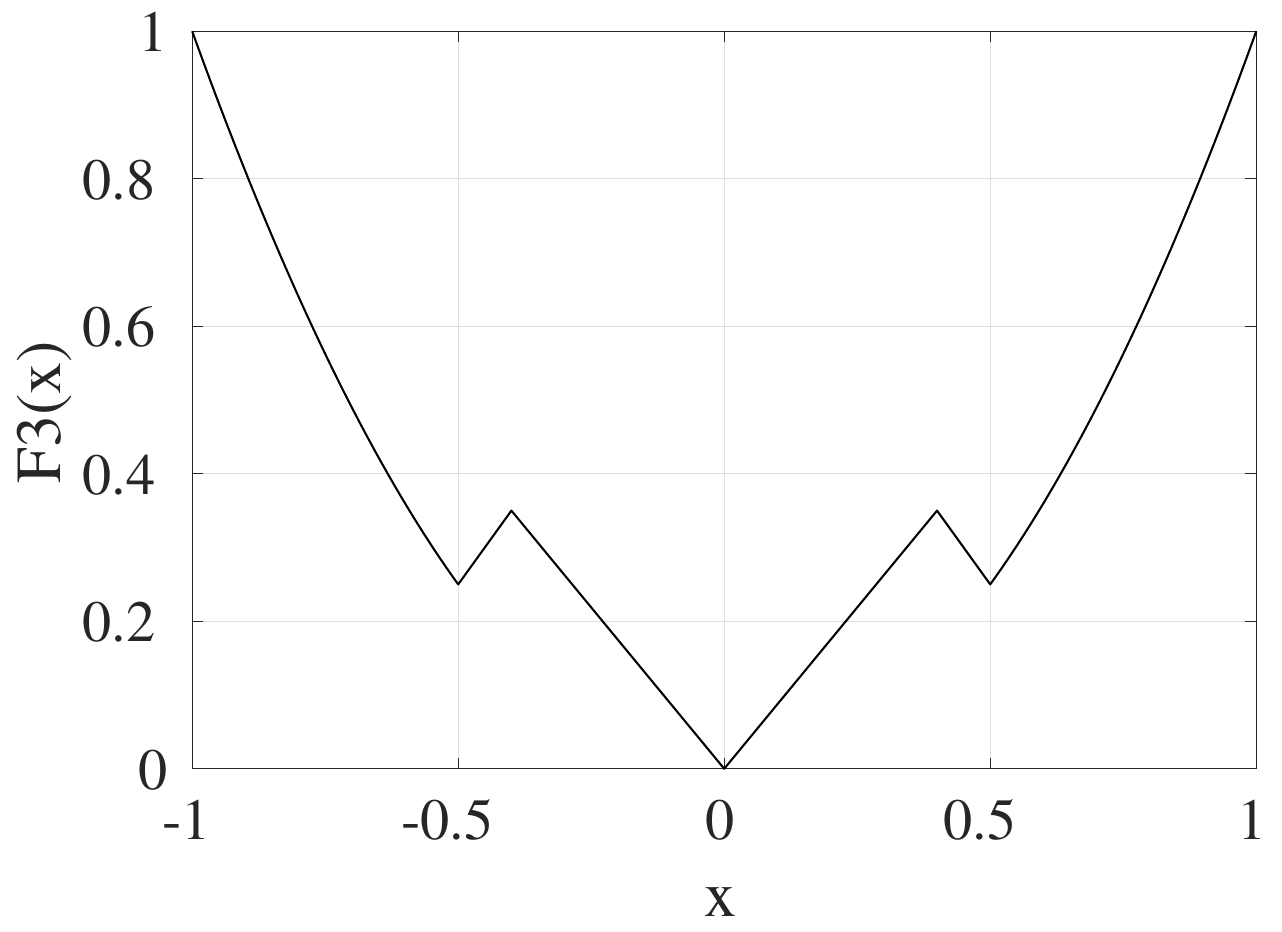}
    \caption{}
    \label{f3}
  \end{subfigure}%
  \begin{subfigure}{.3\textwidth}
    \centering
    \includegraphics[width=1\linewidth]{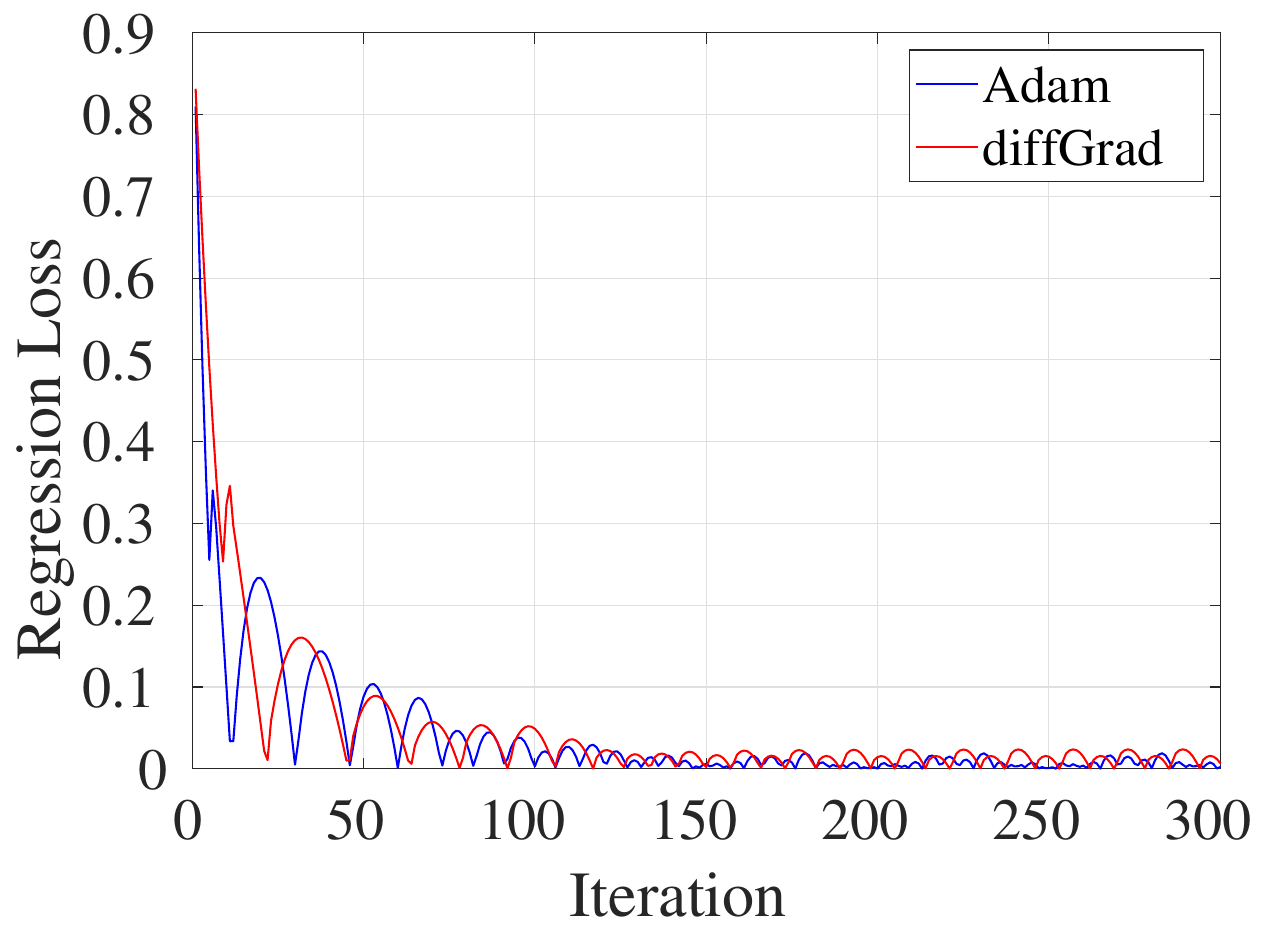}
    \caption{}
    \label{f3_loss}
  \end{subfigure}%
  \begin{subfigure}{.3\textwidth}
    \centering
    \includegraphics[width=1\linewidth]{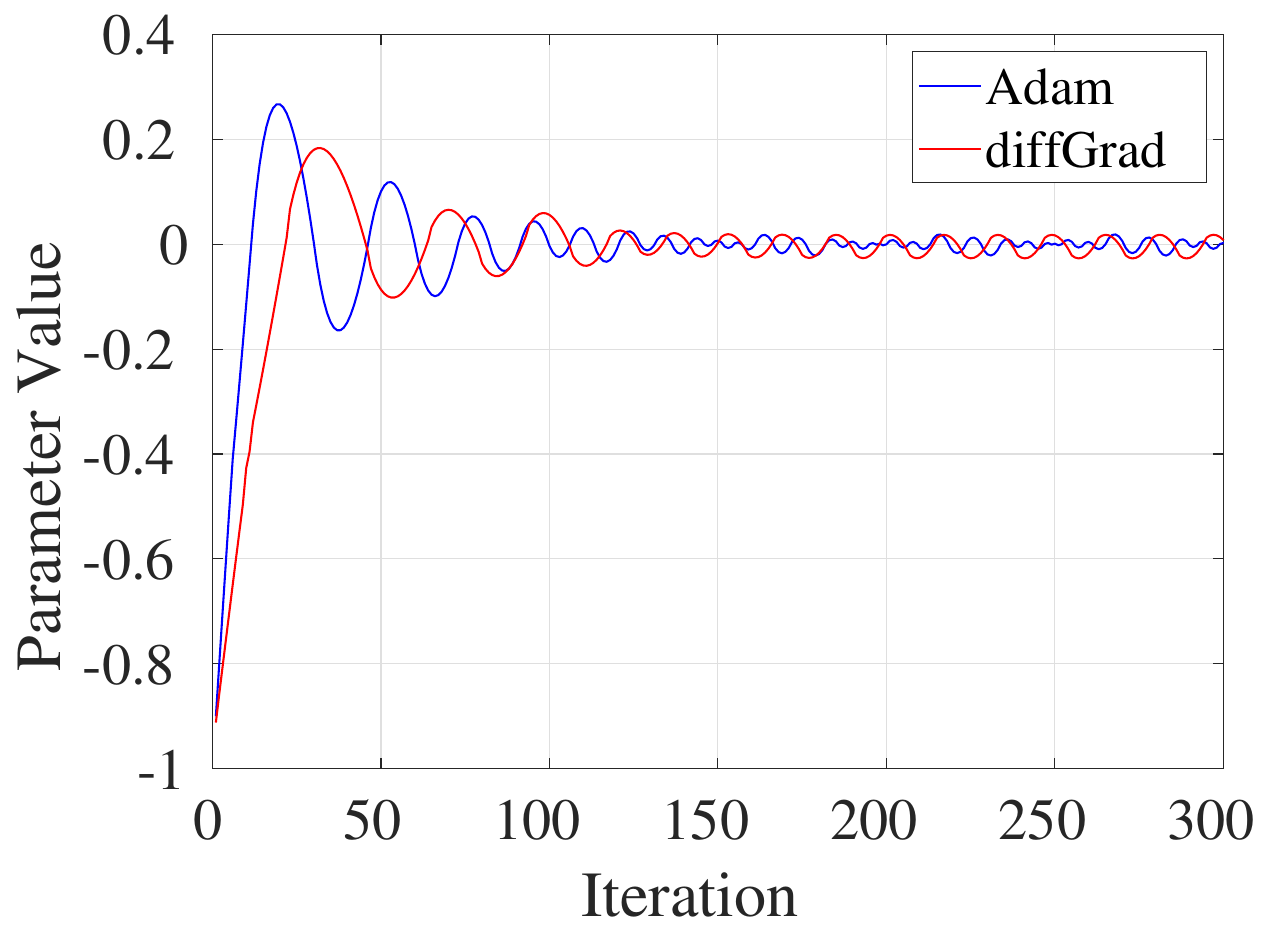}
    \caption{}
    \label{f3_p}
  \end{subfigure}
  \caption{The empirical results comparison among the Adam \cite{Adam} and the proposed diffGrad optimization techniques over three synthetic non-convex functions. (a,d,g) The non-convex synthetic functions $F1$, $F2$, $F3$, respectively, (b,e,h) The regression loss over functions $F1$, $F2$, $F3$, respectively at each iteration, (c,f,i) The parameter value $\theta$ after each iteration during optimization of functions $F1$, $F2$, $F3$, respectively.}
  \label{fig:analytical_results}
\end{figure*}

\section{Empirical Analysis}
In order to justify the purpose of introducing the difference of gradient based friction in diffGrad optimization, we have conducted an empirical analysis in this section. We have modeled the optimization problem as a regression problem over three one-dimensional non-convex functions. We have performed the optimization over these functions by using the widely used Adam \cite{Adam} and the proposed diffGrad optimization methods.

The following non-convex functions, represented by $F1$, $F2$, and $F3$, respectively, are used for this empirical analysis:
\begin{equation}
F1(x)=
\begin{cases}
(x+0.3)^2, & \text{for } x \leq 0 \\ 
(x-0.2)^2+0.05, & \text{for } x > 0 \\
\end{cases}
\end{equation}
\begin{equation}
F2(x)=
\begin{cases}
-40x-35.15, & \text{for } x \leq -0.9 \\ 
x^3+x\sin(8x)+0.85, & \text{for } x > -0.9 \\
\end{cases}
\end{equation}
\begin{equation}
F3(x)=
\begin{cases}
x^2, & \text{for } x \leq -0.5 \\
0.75+x, & \text{for } -0.5 < x \leq -0.4\\
-7x/8, & \text{for } -0.4 < x \leq 0\\
7x/8, & \text{for } 0 < x \leq 0.4\\
0.75-x, & \text{for } 0.4 < x \leq 0.5\\
x^2, & \text{for } 0.5 < x \\
\end{cases}
\end{equation}
where $x$ is the input for this function with $-\infty < x < +\infty$. Functions $F1$, $F2$, and $F3$ are shown in Fig. \ref{f1}, Fig. \ref{f2}, and Fig. \ref{f3}, respectively, for $-1 < x < +1$. 

It can be observed that function $F1$ has one global minimum and one local minimum, whereas functions $F2$ and $F3$ have one global minimum and two local minima. In this experiment, for both Adam \cite{Adam} as well as the proposed diffGrad optimization methods, the following are the hyper-parameter settings: the decay rate for $1^{st}$ moment $(\beta_1)$ is $0.95$; the decay rate for $2^{nd}$ moment $(\beta_2)$ is $0.999$; the learning rate $(\eta)$ is $0.1$ for each iteration; both the $1^{st}$ order moment $(m)$ and $2^{nd}$ order moment $(v)$ are initialized to $0$; and the parameter $\theta$ is initialized to $-1$ in order to show the advantages of the proposed method. In the proposed diffGrad optimization method, the previous gradient value at the $1^{st}$ iteration (i.e., $g_0$) is set to zero. We run Adam \cite{Adam} and the proposed diffGrad optimization for $300$ iterations for all functions. The regression loss as well as parameter value $\theta$ are recorded for both optimization methods at each iteration and analyzed.

Fig. \ref{f1_loss} and Fig. \ref{f1_p} depict the regression loss and the parameter value ($\theta$), respectively, after each iteration for function $F1$. It is discovered from these plots that Adam overshoots the global minimum due to the high moment gained so far. 
It can be observed in Fig. \ref{f1_p} that Adam overshoots the global minimum at $\theta=-0.3$ and becomes stuck in the local minimum at $\theta=0.2$. This problem is addressed by DFC of the proposed diffGrad optimization method, which controls the momentum while reaching towards the global minimum and the overshoot does not occur. Moreover, diffGrad is able to reach a zero loss, as opposed to Adam, which saturates with reasonable loss. The same behavior is also observed for function $F2$, as shown in Fig. \ref{f2_loss} and Fig. \ref{f2_p}.

Both Adam and diffGrad are able to achieve the global minimum for function $F3$, as shown in Fig. \ref{f3_loss} and Fig. \ref{f3_p}. Note that one local minimum is present before the global minimum in function $F3$ (see Fig. \ref{f3}). Both Adam and diffGrad accumulate enough momentum to cross the local minimum. It can be seen in Fig. \ref{f3_p} that Adam oscillates with higher frequency and amplitude around the global minimum, as compared to diffGrad. Thus, better stability is obtained by diffGrad.

\section{Experimental Setup for Classification}
This section presents the setup and settings used in the experiments in terms of the deep architecture used, the hyper-parameter setting, and the applied dataset. 

\subsection{Deep Architecture Used}
The experiments are conducted for an image categorization problem. Convolutional Neural Networks (CNN) are generally used for processing the images. The popular CNN architectures for image categorization problems are AlexNet \cite{AlexNet}, VGGNet \cite{VggNet}, GoogleNet \cite{GoogleNet}, and ResNet \cite{ResNet}. The ResNet based CNN architecture introduced by He et al. \cite{ResNet} is one of the most accurate models which won the  ImageNet Large Scale Visual Recognition Challenge (ILSVRC) \cite{imagenet} in 2015. The ResNet architecture is made with the residual unit. A residual unit considers the learning module as the residual of output w.r.t. the input by implementing a direct connection between input and output of that unit. The residual unit facilitates the training of deeper ResNet architecture which was not feasible in the earlier CNN architectures.
The PyTorch implementation of ResNet, publicly available through GitHub\footnote{\url{https://github.com/kuangliu/pytorch-cifar}}, is used for our experiments. The depth of ResNet is considered as 50 in this paper. Following is the ResNet50 architecture details for the CIFAR datasets: Conv: \{$3\times3$, $64$\}, BatchNorm, Bottleneck(3 times): \{Conv: [$1\times1$, $64$], BatchNorm, Conv: [$3\times3$, $64$], BatchNorm: [$1\times1$, $256$], BatchNorm\}, Bottleneck(4 times): \{Conv: [$1\times1$, $128$], BatchNorm, Conv: [$3\times3$, $128$], BatchNorm, Conv: [$1\times1$, $512$], BatchNorm\}, Bottleneck(6 times): \{Conv: [$1\times1$, $256$], BatchNorm, Conv: [$3\times3$, $256$], BatchNorm, Conv: [$1\times1$, $1024$], BatchNorm\}, Bottleneck(3 times): \{Conv: [$1\times1$, $512$], BatchNorm, Conv: [$3\times3$, $512$], BatchNorm, Conv: [$1\times1$, $2048$], BatchNorm\}, AveragePooling: \{$4\times4$\}. For more information regarding the used ResNet50 architecture, the above mentioned GitHub page may be consulted.

\subsection{Hyper-Parameter Setting}
The hyper-parameter settings in the experiments are illustrated in this paragraph. Stochastic Gradient Descent with momentum (SGDM) is used as the optimization technique. The batch sizes of 32, 64 and 128 are opted. The number of epochs is 100 with a learning rate of $10^{-3}$ for the first 80 epochs and $10^{-4}$ for the last 20 epochs. For all the optimizers, the default settings of PyTorch are used, excluding the moment coefficient for SGDM, which is set to 0.9. 

\begin{table}[t]
\caption{The comparison results in terms of the `Validation Classification Accuracy' over the CIFAR10 and CIFAR100 databases among SGDM, AdaGrad, AdaDelta, RMSProp, AMSGrad, Adam and proposed diffGrad optimization methods. The comparison is made for the batch sizes ($N_b$) of 32, 64 and 128. The best results among different optimization techniques are highlighted in bold.}
\label{results:main}
\begin{center}
\scalebox{1}{
\begin{tabular}{m{1.1cm}| m{.75cm} m{.75cm} m{.8cm}| m{.75cm} m{.75cm} m{.8cm}}
\hline
\multirow{2}{*}{\textbf{Optimizer}} & \multicolumn{3}{c|}{CIFAR10 Database} & \multicolumn{3}{c}{CIFAR100 Database}\\
\cline{2-7}
& $N_b$=32 & $N_b$=64 & $N_b$=128 & $N_b$=32 & $N_b$=64 & $N_b$=128 \\
 \hline
SGDM & 92.95 & 92.07 & 90.15 & 73.3 & 70.37 & 67.25\\
AdaGrad & 92.04 & 92.01 & 91.3 & 71.01 & 70.8 & 68.38\\
AdaDelta & 93.66 & 93.48 & 93.54 & 73.46 & 74.09 & 74.12\\
RMSProp & 92.82 & 92.32 & 92.26 & 65.31 & 66.8 & 62.65\\
AMSGrad & 93.73 & 93.4 & 93.51 & 73.06 & 72.45 & 72.86\\
Adam & 93.78 & 93.81 & 93.72 & 71.82 & 73.31 & 73.72\\
\hline
diffGrad & \textbf{94.24} & \textbf{94.24} & \textbf{94.27} & \textbf{75.63} & \textbf{76.18} & \textbf{75.57}\\
\hline
\end{tabular}
}
\end{center}
\end{table}

\begin{table}[t]
\caption{The comparison results in terms of the `Validation Classification Accuracy' over the CIFAR10 and CIFAR100 databases among different variants of diffGrad. The comparison is made for the batch sizes ($N_b$) of 32, 64 and 128. The best results among different optimization techniques are highlighted in bold.}
\label{results:self}
\begin{center}
\scalebox{1}{
\begin{tabular}{m{1.1cm}| m{.75cm} m{.75cm} m{.8cm}| m{.75cm} m{.75cm} m{.8cm}}
\hline
\multirow{2}{*}{\textbf{Optimizer}} & \multicolumn{3}{c|}{CIFAR10 Database} & \multicolumn{3}{c}{CIFAR100 Database}\\
\cline{2-7}
& $N_b$=32 & $N_b$=64 & $N_b$=128 & $N_b$=32 & $N_b$=64 & $N_b$=128 \\
 \hline
diffGrad & 94.24 & 94.24 & \textbf{94.27} & \textbf{75.63} & \textbf{76.18} & 75.57\\
diffGrad1 & 94.34 & 94.31 & 94.06 & 75.23 & 75.24 & 75.25 \\
diffGrad2 & 94.08 & 93.92 & 94.09 & 75.37 & 75.5 & \textbf{75.83} \\
diffGrad3 & 94.01 & 94.2 & 93.58 & 75.6 & 76.15 & 75.39 \\
diffGrad4 & 94.06 & \textbf{94.37} & 93.75 & 74.9 & 75.35 & 75.45\\
diffGrad5 & \textbf{94.42} & 94.02 & 94.16 & 75.55 & 75.91 & 75.16\\
\hline
\end{tabular}
}
\end{center}
\end{table}

\subsection{Dataset Used}
In order to conduct the image categorization experiments, the CIFAR10 and CIFAR100 datasets\footnote{\url{https://www.cs.toronto.edu/~kriz/cifar.html}} \cite{cifar} are used in this paper. Both the CIFAR10 and the CIFAR100 datasets consist of the same 60000 images, including 50000 images for training and 10000 images for validation. In the CIFAR10 dataset, all images are divided into 10 categories including `airplane', `automobile', `bird', `cat', `deer', `dog', `frog', `horse', `ship', and `truck'. Whereas, in the CIFAR100 dataset, the same set of images is partitioned into 100 categories. The dimension of all images is $32\times32\times3$. 
Note that the images are pre-processed to make the RGB values zero-centered with unit standard deviation across each color channel. The data augmentation is done only over the training images by the process of flipping and cropping. The images are randomly flipped horizontally (i.e., with respect to the vertical axis) with a probability of 0.5. In case of cropping, at first the images are scaled to $40\times40\times3$ size by zero padding. Then, the images of size $32\times32\times3$ are cropped randomly from the upscaled images.

\section{Classification Experiments and Analysis}
The image categorization experiments over the CIFAR10 and CIFAR100 datasets are conducted to test the performance improvement of the proposed diffGrad gradient descent optimization method. The ResNet50 is used to demonstrate the suitability of diffGrad for the CNN model. In this section, at first the results of the proposed diffGrad method are presented, then the results are compared with other state-of-the-art optimization methods, and finally the stability of the proposed diffGrad optimization is tested over different activation functions. The results are computed in terms of the average top-1 validation classification accuracy.

\subsection{Validation Results Comparison}
The validation classification accuracy due to the diffGrad method is compared with state-of-the-art optimization techniques such as SGDM \cite{SGDM}, AdaGrad \cite{AdaGrad}, AdaDelta \cite{AdaDelta}, RMSProp \cite{RMSProp}, AMSGrad \cite{AMSGrad}, and ADAM \cite{Adam}. Table \ref{results:main} depicts the validation classification accuracy for different optimization techniques. The results are presented over the CIFAR10 and CIFAR100 datasets. The best results among the different optimization techniques are highlighted in bold. It is observed from Table \ref{results:main} that the proposed diffGrad optimization technique outperforms all other optimization techniques over both the CIFAR10 and the CIFAR100 datasets for all tested batch sizes of 32, 64, and 128, respectively. The proposed diffGrad method utilizes all the positive characteristics of Adam. Moreover, the effect of the proposed difference of gradient based friction technique prevents the network from noisy oscillation near the minimum solution. It leads to more accurate results as compared to other optimization techniques. 

\subsection{Experiments with diffGrad Variants}
In this section, we modify the diffGrad Friction Coefficient (DFC) of diffGrad and analyze the performance over the CIFAR10 dataset. The DFC is basically given as $DFC=1/(1+e^{-(|g_{t-1}-g_{t}|)})$ with the range $DFC \in [0.5,1]$. We modify the DFC by removing the absolute value, which is given as $DFC1=1/(1+e^{-(g_{t-1}-g_{t})})$. Note that $DFC1 \in [0,1]$. We also generate another version called DFC2, given by $DFC2=9/(1+e^{-(0.5|g_{t-1}-g_{t}|})-4)$ with the range $DFC2 \in [0.5,5]$. We also use the mean ($\mu$) and standard deviation ($\nu$) of absolute gradients of the batch with DFC and consider the following three more scenarios: (a) $DFC3=1/(1+e^{-(\nu|g_{t-1}-g_{t}|-\mu)})$ with the range $DFC3 \in [0.5,1]$, (b) $DFC4=1/(1+e^{-(\nu^2|g_{t-1}-g_{t}|-\mu)})$ with the range $DFC4 \in [0.5,1]$, and (c) $DFC5=1/(1+e^{-(\sqrt{\nu}|g_{t-1}-g_{t}|-\mu)})$ with the range $DFC5 \in [0.5,1]$. We define numbered diffGrad accordingly, i.e., diffGrad1 using DFC1 and so on. The validation classification accuracy for diffGrad using DFC, diffGrad1 using DFC1, diffGrad2 using DFC2, diffGrad3 using DFC3, diffGrad4 using DFC4, and diffGrad5 using DFC5 are presented in Table \ref{results:self}. The results are computed over the CIFAR10 and CIFAR100 datasets with batch sizes 32, 64, and 128. It can be noticed from this result that the original diffGrad performs better over CIFAR10 dataset for high batch size and over CIFAR100 dataset for small batch sizes. The mean and standard deviation based diffGrad variants such as DFC4 and DFC5 perform better over CIFAR10 dataset for 64 and 32 bach sizes, respectively. It is due to the fact that the mean and standard deviation of the absolute gradient tends to be more accurate for higher batch size. From Table \ref{results:main} and Table \ref{results:self}, it is clear that original diffGrad as well as its mean and standard deviation based variants show the promising performance.

\begin{table}[t]
\caption{The performance of the proposed diffGrad optimization technique with ResNet50 model for different activation functions, namely ReLU, Leaky ReLU (LReLU), ELU, and SELU. The Validation Classification Accuracy over CIFAR10 is reported for the batch sizes ($N_b$) of 32, 64 and 128. The best results among different activation's are highlighted in bold.}
\label{results:activation}
\begin{center}
\scalebox{1}{
\begin{tabular}{c| c c c}
\hline
\multirow{2}{*}{\textbf{Optimizer}} & \multicolumn{3}{c}{CIFAR10 Database}\\
\cline{2-4}
& $N_b$=32 & $N_b$=64 & $N_b$=128\\
 \hline
ResNet50(ReLU) & 94.24 & \textbf{94.24} & 94.27\\
ResNet50(LReLU) & 94.3 & 94.2 & \textbf{94.3}\\
ResNet50(ELU) & \textbf{94.48} & 94.2 & 94.24\\
ResNet50(SELU) & 94.03 & 93.62 & 93.96\\
\hline
\end{tabular}
}
\end{center}
\end{table}

\subsection{Performance Analysis with Activation Functions}
In other experiments, the Rectified Linear Unit (ReLU) \cite{AlexNet} is used as the default activation function in the framework of ResNet \cite{ResNet}. In this experiment, we have considered four activation functions, namely, ReLU \cite{AlexNet}, Leaky ReLU (LReLU) \cite{lrelu}, Exponential Linear Unit (ELU) \cite{elu}, and Scaled ELU (SELU) \cite{selu}. The performance of the proposed diffGrad optimization method is computed over the CIFAR10 dataset for different activation functions. We have used the same experimental setup as for the ResNet50 model by replacing all the activation functions with ReLU, LReLU, ELU, and SELU, one by one. The rest of the experimental setup is same as the setup of previous experiments. The leaky factor in LReLU is considered as 0.1. The validation classification accuracy is compared in Table \ref{results:activation}. It is clear from this result that the ELU, ReLU and LReLU outperform others for the batch sizes of 32. 64 and 128, respectively.

\section{Conclusion}
In this paper, a new stochastic gradient descent optimization method diffGrad is proposed. diffGrad incorporates the difference of gradients of current and immediate past iteration (i.e., short term gradient change information) with Adam optimization techniques to control the learning rate based on the optimization stage. The proposed diffGrad allows a high learning rate if the gradient change is more (i.e., the optimization is far from the optimum solution), and a low learning rate if the gradient changes minimally (i.e., the optimization is near to the optimum solution). The local optima and saddle point scenarios are handled by the moment gained due to past gradients. The regret bound analysis provides a guarantee of convergence. An empirical analysis over three synthetic, non-convex functions reveals that the proposed diffGrad optimization method controls the update step in order to avoid the overshooting of global minimum and oscillation around global minimum. The proposed diffGrad optimization method is also tested with the ResNet50 model for an image categorization task over the CIFAR10 and CIFAR100 datasets. The results are compared with the state-of-the-art SGD optimization techniques such as SGDM, AdaGrad, AdaDelta, RMSProp, AMSGrad, and Adam. It is observed that the diffGrad outperforms all other optimizers. Moreover, the mean and standard deviation of the absolute gradient of a batch can be used with diffGrad for better result over CIFAR10 dataset.

\section*{Acknowledgments}
The authors would like to thank NVIDIA Corporation for the support of GeForce Titan X Pascal GPU donated to Computer Vision Group, IIIT Sri City. The authors would also like to thank the anonymous Associate Editor and Reviewers for their valuable comments to improve the quality of the paper. The authors would also like to thank Dr. Shrijita for editing the paper.

\ifCLASSOPTIONcaptionsoff
  \newpage
\fi

{\small
\bibliographystyle{IEEEtran}
\bibliography{Reference}
}

\section*{Appendix}
\subsection{Convergence Proof}
\begin{theorem}
\textit{Let the bounded gradients for function $f_t$ (i.e., $||g_{t,\theta}||_2 \leq G$ and $||g_{t,\theta}||_{\infty} \leq G_{\infty}$) for all $\theta \in R^d$. Also assume that diffGrad produces the bounded distance between any $\theta_t$ (i.e., $||\theta_n-\theta_m||_2 \leq D$ and $||\theta_n-\theta_m||_\infty \leq D_\infty$ for any $m,n\in\{1,...,T\}$). Let $\gamma \triangleq \frac{\beta_1^2}{\sqrt{\beta_2}}$, $\beta_1,\beta_2 \in [0,1)$ satisfy $\frac{\beta_1^2}{\sqrt{\beta_2}} < 1$, $\alpha_t=\frac{\alpha}{\sqrt{t}}$, and $\beta_{1,t}=\beta_1\lambda^{t-1},\lambda \in (0,1)$ with $\lambda$ is typically close to $1$, e.g $1-10^{-8}$. For all $T \geq 1$, the proposed diffGrad optimizer shows the following guarantee:} 
\begin{dmath}
R(T) \leq \frac{D^2}{2\alpha(1-\beta_1)}\sum_{i=1}^{d}{(1+e^{-|g_{1,i}|})\sqrt{T\hat{v}_{T,i}}} 
+ \frac{\alpha(1+\beta_1) G_\infty}{(1-\beta_1)\sqrt{1-\beta_2}(1-\gamma)^2}\sum_{i=1}^{d}{||g_{1:T,i}||_2} 
+ \sum_{i=1}^{d}{\frac{D_{\infty}^{2}G_{\infty}\sqrt{1-\beta_2}}{2\alpha (1-\beta_1)(1-\lambda)^2}}
\end{dmath}

\begin{proof}[Proof]
Using Lemma 10.2 of Adam \cite{Adam}, we can write as
$$
f_t(\theta_t)-f_t(\theta^*) \leq g_t^T(\theta_t-\theta^*) = \sum_{i=1}^{d}{g_{t,i}(\theta_{t,i}-\theta_{,i}^*)}
$$
We can write following from the diffGrad update rule described in Eq. (\ref{diffgrad_update_rule}), ignoring $\epsilon0$, 

\begin{dmath}
\theta_{t+1} =\theta_t-\frac{\alpha_t \xi_t \hat{m}_t}{\sqrt[]{\hat{v}_{t}}}
 =\theta_t-\frac{\alpha_t \xi_t}{(1-\beta_1^t)} \Big(\frac{\beta_{1,t}}{\sqrt[]{\hat{v}_{t}}}m_{t-1} + \frac{(1-\beta_{1,t})}{\sqrt[]{\hat{v}_{t}}}g_t\Big)
\end{dmath}
where $\beta_{1,t}$ is the $1^{st}$ order moment at $t^{th}$ iteration and $\beta_1^t$ is the $t^{th}$ power of initial $1^{st}$ order moment. \\
For $i^{th}$ dimension of parameter vector $\theta_t \in R^d$, we can write
\begin{dmath}
(\theta_{t+1,i}-\theta_{,i}^*)^2=(\theta_{t,i}-\theta_{,i}^*)^2-\frac{2\alpha_t \xi_{t,i}}{1-\beta_1^t}
\Big(\frac{\beta_{1,t}}{\sqrt[]{\hat{v}_{t,i}}}m_{t-1,i} + \frac{(1-\beta_{1,t})}{\sqrt[]{\hat{v}_{t,i}}}g_{t,i}\Big)(\theta_{t,i}-\theta_{,i}^*)
+\alpha_t^2 \xi_{t,i}^2 (\frac{\hat{m}_{t,i}}{\hat{v}_{t,i}})^2
\end{dmath}
The above equation can be reordered as
\begin{dmath}
g_{t,i}(\theta_{t,i}-\theta_{,i}^*)=\frac{(1-\beta_1^t)\sqrt{\hat{v}_{t,i}}}{2\alpha_t\xi_{t,i}(1-\beta_{1,t})}
\Big((\theta_{t,i}-\theta_{,i}^*)^2-(\theta_{t+1,i}-\theta_{,i}^*)^2\Big)
+\frac{\beta_{1,t}}{1-\beta_{1,t}}(\theta_{,i}^*-\theta_{t,i})m_{t-1,i}
+\frac{\alpha_t(1-\beta_1^t)\xi_{t,i}}{2(1-\beta_{1,t})}\frac{(\hat{m}_{t,i})^2}{\sqrt{\hat{v}_{t,i}}}.
\end{dmath}
Further, it can be written as
\begin{dmath}
g_{t,i}(\theta_{t,i}-\theta_{,i}^*) =\frac{(1-\beta_1^t)\sqrt{\hat{v}_{t,i}}}{2\alpha_t\xi_{t,i}(1-\beta_{1,t})}
\Big((\theta_{t,i}-\theta_{,i}^*)^2-(\theta_{t+1,i}-\theta_{,i}^*)^2\Big)
 +\sqrt{\frac{\beta_{1,t}}{\alpha_{t-1}(1-\beta_{1,t})}(\theta_{,i}^*-\theta_{t,i})^2\sqrt{\hat{v}_{t-1,i}}} \sqrt{\frac{\beta_{1,t}\alpha_{t-1}(m_{t-1,i})^2}{(1-\beta_{1,t})\sqrt{\hat{v}_{t-1,i}}}}
+\frac{\alpha_t(1-\beta_1^t)\xi_{t,i}}{2(1-\beta_{1,t})}\frac{(\hat{m}_{t,i})^2}{\sqrt{\hat{v}_{t,i}}}
\end{dmath}
Based on Young's inequality, $ab \leq a^2/2+b^2/2$ and fact that $\beta_{1,t} \leq \beta_1$, the above equation can be reordered as 
\begin{dmath}
g_{t,i}(\theta_{t,i}-\theta_{,i}^*) \leq \frac{1}{2\alpha_t\xi_{t,i}(1-\beta_1)}
\Big((\theta_{t,i}-\theta_{,i}^*)^2-(\theta_{t+1,i}-\theta_{,i}^*)^2\Big)\sqrt{\hat{v}_{t,i}}
 +\frac{\beta_{1,t}}{2\alpha_{t-1}(1-\beta_{1,t})}(\theta_{,i}^*-\theta_{t,i})^2\sqrt{\hat{v}_{t-1,i}} + \frac{\beta_1\alpha_{t-1}(m_{t-1,i})^2}{2(1-\beta_1)\sqrt{\hat{v}_{t-1,i}}}
+\frac{\alpha_t\xi_{t,i}}{2(1-\beta_1)}\frac{(\hat{m}_{t,i})^2}{\sqrt{\hat{v}_{t,i}}}
\end{dmath}
From Eq. (\ref{absig}) and Fig. \ref{fig:sigmoid}, it is clear that $0.5 \leq \xi_{t,i} \leq 1$. So, $\xi_{t,i}$ can be removed from last term of above equation and it still satisfy the inequality. Then,
\begin{dmath}
g_{t,i}(\theta_{t,i}-\theta_{,i}^*) \leq \frac{1}{2\alpha_t\xi_{t,i}(1-\beta_1)}
\Big((\theta_{t,i}-\theta_{,i}^*)^2-(\theta_{t+1,i}-\theta_{,i}^*)^2\Big)\sqrt{\hat{v}_{t,i}}
 +\frac{\beta_{1,t}}{2\alpha_{t-1}(1-\beta_{1,t})}(\theta_{,i}^*-\theta_{t,i})^2\sqrt{\hat{v}_{t-1,i}} + \frac{\beta_1\alpha_{t-1}(m_{t-1,i})^2}{2(1-\beta_1)\sqrt{\hat{v}_{t-1,i}}}
+\frac{\alpha_t}{2(1-\beta_1)}\frac{(\hat{m}_{t,i})^2}{\sqrt{\hat{v}_{t,i}}}
\end{dmath}
We use the Lemma 10.4 of Aadm \cite{Adam} and derive the regret bound by aggregating it across all the dimensions for $i\in \{1,\dots,d\}$ and all the sequence of convex functions for $t\in \{1,\dots,T\}$ in the upper bound of $f_t(\theta_t)-f_t(\theta^*)$ as
\begin{dmath}
R(T) \leq \sum_{i=1}^{d}{\frac{1}{2\alpha_1\xi_{1,i}(1-\beta_1)}} (\theta_{1,i}-\theta_{,i}^*)^2\sqrt{\hat{v}_{1,i}} + \sum_{i=1}^{d}{\sum_{t=2}^{T}{\frac{1}{2(1-\beta_1)}} (\theta_{t,i}-\theta_{,i}^*)^2(\frac{\sqrt{\hat{v}_{t,i}}}{\alpha_t\xi_{t,i}}-\frac{\sqrt{\hat{v}_{t-1,i}}}{\alpha_{t-1}\xi_{t-1,i}})}
 + \frac{\beta_1\alpha G_\infty}{(1-\beta_1)\sqrt{1-\beta_2}(1-\gamma)^2}\sum_{i=1}^{d}{||g_{1:T,i}||_2}
+ \frac{\alpha G_\infty}{(1-\beta_1)\sqrt{1-\beta_2}(1-\gamma)^2}\sum_{i=1}^{d}{||g_{1:T,i}||_2}
 + \sum_{i=1}^{d}{\sum_{t=1}^{T}{\frac{\beta_{1,t}}{2\alpha_{t}(1-\beta_{1,t})}(\theta_{,i}^*-\theta_{t,i})^2\sqrt{\hat{v}_{t,i}}}}
\end{dmath}
By utilizing the assumptions that $\alpha=\alpha_t\sqrt{t}$, $||\theta_t-\theta^*||_2 \leq D$ and $||\theta_m-\theta_n||_{\infty} \leq D_{\infty}$, we can write as
\begin{dmath}
R(T) \leq \frac{D^2}{2\alpha(1-\beta_1)}\sum_{i=1}^{d}{\frac{\sqrt{T\hat{v}_{T,i}}}{\xi_{1,i}}} 
+ \frac{\alpha(1+\beta_1) G_\infty}{(1-\beta_1)\sqrt{1-\beta_2}(1-\gamma)^2}\sum_{i=1}^{d}{||g_{1:T,i}||_2}
+ \frac{D_{\infty}^{2}}{2\alpha}\sum_{i=1}^{d}{\sum_{t=1}^{t}{\frac{\beta_{1,t}}{(1-\beta_{1,t})}\sqrt{t\hat{v}_{t,i}}}}
  \leq \frac{D^2}{2\alpha(1-\beta_1)}\sum_{i=1}^{d}{\frac{\sqrt{T\hat{v}_{T,i}}}{\xi_{1,i}}} 
+ \frac{\alpha(1+\beta_1) G_\infty}{(1-\beta_1)\sqrt{1-\beta_2}(1-\gamma)^2}\sum_{i=1}^{d}{||g_{1:T,i}||_2} 
+ \frac{D_{\infty}^{2}G_{\infty}\sqrt{1-\beta_2}}{2\alpha}\sum_{i=1}^{d}{\sum_{t=1}^{t}{\frac{\beta_{1,t}}{(1-\beta_{1,t})}\sqrt{t}}}
\end{dmath}
It is shown in Adam \cite{Adam} that $\sum_{t=1}^{t}{\frac{\beta_{1,t}}{(1-\beta_{1,t})}\sqrt{t}} \leq \frac{1}{(1-\beta_1)(1-\gamma)^2}$. Thus, the regret bound can be written as
\begin{dmath}
R(T) \leq \frac{D^2}{2\alpha(1-\beta_1)}\sum_{i=1}^{d}{\frac{\sqrt{T\hat{v}_{T,i}}}{\xi_{1,i}}} 
+ \frac{\alpha(1+\beta_1) G_\infty}{(1-\beta_1)\sqrt{1-\beta_2}(1-\gamma)^2}\sum_{i=1}^{d}{||g_{1:T,i}||_2} 
+ \sum_{i=1}^{d}{\frac{D_{\infty}^{2}G_{\infty}\sqrt{1-\beta_2}}{2\alpha (1-\beta_1)(1-\lambda)^2}}
\end{dmath}
We know $\xi_{1,i}=1/(1+e^{-|g_{0,i}-g_{1,i}|})=1/(1+e^{-|g_{1,i}|})$ as $g_{0,i}=0$. Therefore, the regret bound for diffGrad is as follows:
\begin{dmath}
R(T) \leq \frac{D^2}{2\alpha(1-\beta_1)}\sum_{i=1}^{d}{(1+e^{-|g_{1,i}|})\sqrt{T\hat{v}_{T,i}}} 
+ \frac{\alpha(1+\beta_1) G_\infty}{(1-\beta_1)\sqrt{1-\beta_2}(1-\gamma)^2}\sum_{i=1}^{d}{||g_{1:T,i}||_2} 
+ \sum_{i=1}^{d}{\frac{D_{\infty}^{2}G_{\infty}\sqrt{1-\beta_2}}{2\alpha (1-\beta_1)(1-\lambda)^2}}
\end{dmath}
\end{proof}
\end{theorem}

\begin{IEEEbiography}[{\includegraphics[width=1in,height=1.25in,clip,keepaspectratio]{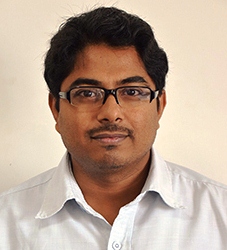}}]{Shiv Ram Dubey}
has been with the Indian Institute of Information Technology (IIIT), Sri City since June 2016, where he is currently the Assistant Professor of Computer Science and Engineering. He received the Ph.D. degree in Computer Vision and Image Processing from Indian Institute of Information Technology, Allahabad (IIIT Allahabad) in 2016. Before that, from August 2012-Feb 2013, he was a Project Officer in the Computer Science and Engineering Department at Indian Institute of Technology, Madras (IIT Madras). 
He was a recipient of several awards including the Indo-Taiwan Joint Research Grant from DST/GITA, Govt. of India, Best PhD Award in PhD Symposium, IEEE-CICT2017 at IIITM Gwalior, Early Career Research Award from SERB, Govt. of India and NVIDIA GPU Grant Award Twice from NVIDIA. He received Outstanding Certificate of Reviewing Award from Information Fusion, Elsevier in 2018. He also received the Best Paper Award in IEEE UPCON 2015, a prestigious conference of IEEE UP Section. 
His research interest includes Computer Vision, Deep Learning, Image Processing, Biometrics, Medical Imaging, Convolutional Neural Networks, Image Feature Description, Content Based Image Retrieval, Image-to-Image Transformation, Face Detection and Recognition, Facial Expression Recognition, Texture and Hyperspectral Image Analysis.  
\end{IEEEbiography}

\begin{IEEEbiography}[{\includegraphics[width=1in,height=1.25in,clip,keepaspectratio]{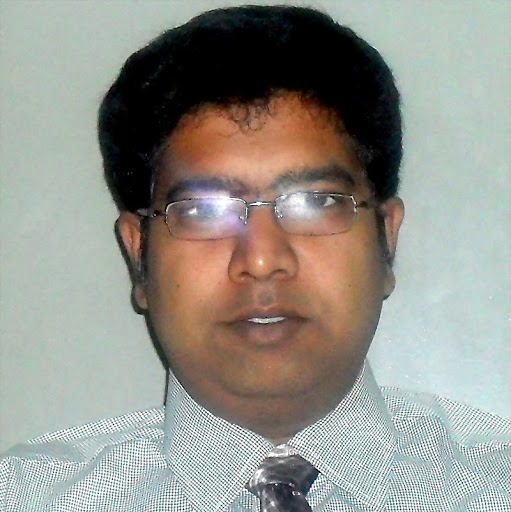}}]{Soumendu Chakraborty}
received his Bachelor of Engineering (B.E.) in Information Technology from University Institute of Technology, University of Burdwan, India in 2005. He did his M.Tech. in Computer Science and Engineering from GLA University, Mathura, India in 2013. He received his Ph.D. from Indian Institute of Information Technology, Allahabad, U.P., India in the year 2018.  He has 12 years of teaching and research experience. Presently, he is working as an Assistant Professor in Indian Institute of Information Technology, Lucknow, India. His research interests include Computer Vision, Machine Learning, image processing, biometric systems, image stegnography and pattern recognition. 
\end{IEEEbiography}

\begin{IEEEbiography}[{\includegraphics[width=1in,height=1.25in,clip,keepaspectratio]{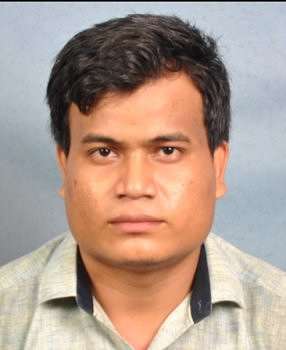}}]{Swalpa Kumar Roy}
received both his Bachelor and Master degree in Computer Science \& Engineering from West Bengal University of Technology,
Kolkata, India, in 2012, and Indian Institute of Engineering Science and
Technology, Shibpur, Howrah, India, in 2015, respectively. He is pursuing the Ph.D. degree from under Computer Vision and Pattern Recognition Unit, Indian Statistical Institute, Kolkata, India, where he also worked as a Project Linked Person from July 2015 to March 2016. He is currently working as an Assistant Professor in the Department of Computer Science \& Engineering, Jalpaiguri Government Engineering College, Jalpaiguri, West Bengal, India. His research interests include Computer Vision, Deep Learning, Remote Sensing, Texture Feature Description, and Fractal Image
Coding.\end{IEEEbiography}

\begin{IEEEbiography}[{\includegraphics[width=1in,height=1.25in,clip,keepaspectratio]{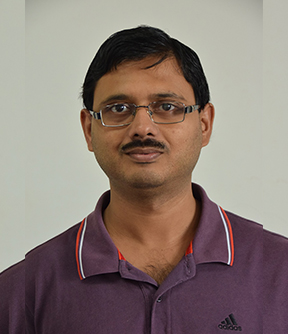}}]{Snehasis Mukherjee}
has obtained his PhD in Computer Science from the Indian Statistical Institute in 2012.
Before doctoral study, he has completed his Bachelors degree in Mathematics from the University of Calcutta
and Masters degree in Computer Applications from the Vidyasagar University. He did his Post Doctoral
Research works at the National Institute of Standards and Technology (NIST), Gaithersburg, Maryland, USA.
Currently he is working as an Assistant Professor in the Indian Institute of Information Technology Chittoor,
SriCity (IIIT Chittoor, SriCity). He has written several peer-reviewed research papers (in reputed journals and
conferences). His research area includes Computer Vision, Machine Learning, Image and Video Processing.
\end{IEEEbiography}

\begin{IEEEbiography}[{\includegraphics[width=1in,height=1.25in,clip,keepaspectratio]{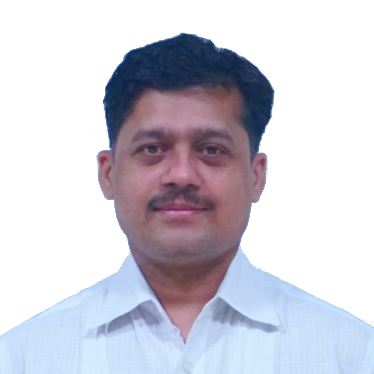}}]{Satish Kumar Singh}
is an Associate Professor 
with the Indian Institute of Information Technology Allahabad, Prayagraj, India. He received the Ph.D., M.Tech., and B.Tech. degrees in 2010, 2005, and 2003, respectively. He has over 13 years of experience in academic and research institutions. He has authored over 45 publications in reputed international journal and conference proceedings. He is a member of 
various professional societies, like the Institution of Electronics and Telecommunication Engineers. He is an Executive Committee Member of the IEEE Uttar Pradesh Section from 2014. Currently he is serving as secretary Signal Processing Society Chapter, and Uttar Pradesh Section as well. He is serving as an Editorial Board Member and reviewer for many international journals. His current research interests are in the areas of digital image processing, pattern recognition, multimedia data indexing and retrieval, watermarking and biometrics.
\end{IEEEbiography}

\begin{IEEEbiography}[{\includegraphics[width=1in,height=1.25in,clip,keepaspectratio]{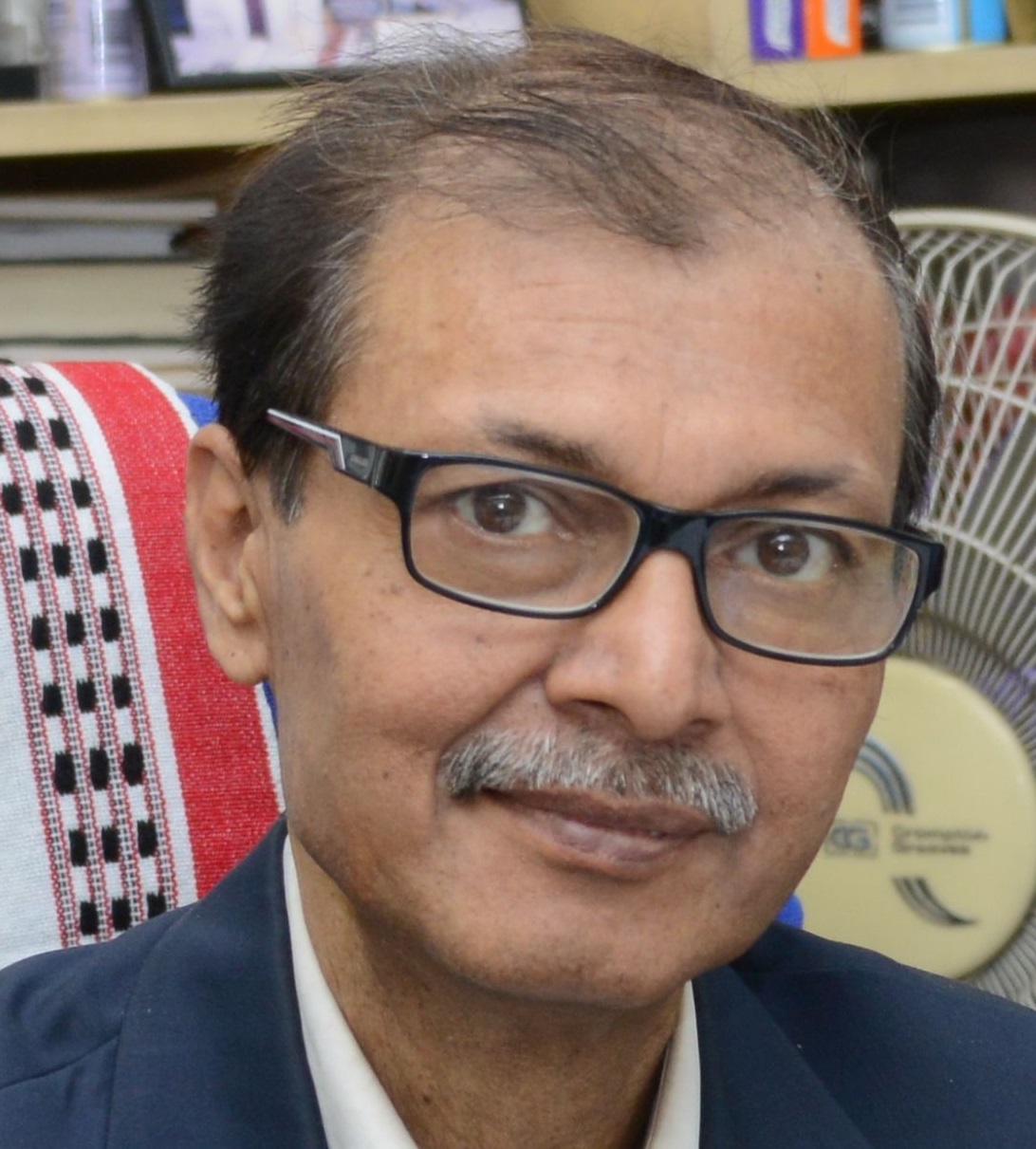}}]{Bidyut Baran Chaudhuri}
received the Ph.D. degree from IIT Kanpur, in 1980. He was a Leverhulme Postdoctoral Fellow with Queen’s University, U.K., from 1981 to 1982. He joined the Indian Statistical Institute, in 1978, where he worked as an INAE Distinguished Professor and a J C Bose Fellow at Computer Vision and Pattern Recognition Unit of Indian Statistical Institute. He is now affiliated to Techno India University, Kolkata as Pro-Vice Chancellor (Academic). His research interests include Pattern Recognition, Image Processing, Computer Vision, Natural Language Processing (NLP), Signal processing, Digital Document Processing, Deep learning etc. He pioneered the first workable OCR system for printed Indian scripts Bangla, Assamese and Devnagari. He also developed computerized \textit{Bharati Braille system} with speech synthesizer and has done statistical analysis of Indian language.  He has published about 425 research papers in international journals and conference proceedings. Also, he has authored/edited seven books in these fields.
Prof Chaudhuri received Leverhulme fellowship award, Sir J. C. Bose Memorial Award, M. N. Saha Memorial Award, Homi Bhabha Fellowship, Dr. Vikram Sarabhai Research Award, C. Achuta Menon Award, Homi Bhabha Award: Applied Sciences, Ram Lal Wadhwa Gold Medal, Jawaharlal Nehru Fellowship, J C Bose fellowship, Om Prakash Bhasin Award etc. 
Prof Chaudhuri is the associate editor of three international journals and a fellow of INSA, NASI, INAE, IAPR, The World Academy of Sciences (TWAS) and life fellow of IEEE (2015). He acted as General Chair and Technical Co-chair at various International Conferences.

\end{IEEEbiography}

\end{document}